%% file: main.tex
\documentclass{article}
\pdfoutput=1
\usepackage[numbers]{natbib}

\usepackage[preprint]{neurips_2023}
\usepackage[utf8]{inputenc} 
\usepackage[T1]{fontenc}
\usepackage{hyperref}
\usepackage{url}
\usepackage{booktabs} 
\usepackage{amsfonts}
\usepackage{booktabs} 
\usepackage{nicefrac}
\usepackage{microtype}
\usepackage{xcolor}
\usepackage{tabularx}
\usepackage{graphicx}
\usepackage{algorithm} 
\usepackage{algorithmic} 
\usepackage{caption}
\usepackage{bibunits}
\usepackage{bm}
\usepackage{amsmath}
\usepackage{xspace}
\mathchardef\mhyphen="2D
\usepackage{subcaption}

\newcommand{\ours}{LfVoid\xspace}
\newcommand{\oursfull}{Can Pre-Trained Text-to-Image Models Generate Visual Goals for Reinforcement Learning?}

\title{\oursfull}
\author{
Jialu Gao$^{1*}$, \quad
Kaizhe Hu$^{1,2}\thanks{equal contribution}$, \quad
Guowei Xu$^{1}$, \quad
Huazhe Xu$^{1,2,3}$\\
\vspace{0.1cm}
$^{1}$ Tsinghua University\quad $^{2}$  Shanghai Qi Zhi Institute \quad $^{3}$ Shanghai AI Lab\\
\texttt{gaojialululu@gmail.com,
huazhe\_xu@mail.tsinghua.edu.cn}
\vspace{1.0cm}
}

\begin{document}

\maketitle

\input{00_abstract}
\input{01_introduction}
\input{02_related_works}
\input{03_methods}
\input{04_experiments}

\input{05_conclusion}
\input{07_acknowledgement}

\bibliographystyle{unsrt}
\bibliography{main}

\newpage
\input{06_appendix}

\end{document}

%% file: 00_abstract.tex
\begin{abstract}
Pre-trained text-to-image generative models can produce diverse, semantically rich, and realistic images from natural language descriptions. 
Compared with language, images usually convey information with more details and less ambiguity. In this study, we propose Learning from the Void (\ours), a method that leverages the power of pre-trained text-to-image models and advanced image editing techniques to guide robot learning. Given natural language instructions, \ours can edit the original observations to obtain goal images, such as ``wiping'' a stain off a table. Subsequently, \ours trains an ensembled goal discriminator on the generated image to provide reward signals for a reinforcement learning agent, guiding it to achieve the goal. The ability of \ours to learn with zero in-domain training on expert demonstrations or true goal observations (the void) is attributed to the utilization of knowledge from web-scale generative models.
We evaluate \ours across three simulated tasks and validate its feasibility in the corresponding real-world scenarios. In addition, we offer insights into the key considerations for the effective integration of visual generative models into robot learning workflows. We posit that our work represents an initial step towards the broader application of pre-trained visual generative models in the robotics field. Our project page: \href{https://lfvoid-rl.github.io/}{\texttt{LfVoid.github.io}}.

\end{abstract}

%% file: 01_introduction.tex
\section{Introduction}

What is the simplest way to provide guidance or goals to a robot? This question is answered multiple times: a set of expert demonstrations~\cite{papagiannis2023imitation, dadashi2020primal, kostrikov2018discriminator, haldar2023watch}, goal images~\cite{fu2018variational,singh2019end,eysenbach2021replacing}, or natural language instructions~\cite{ahn2022can,brohan2022rt}. However, these answers either require a laborious and sometimes prohibitive effort to collect data, or contain ambiguity across modalities. The desire to alleviate the challenges begs the question: can we generate goal images from natural language instructions directly, without physically achieving the goals for robots? 

Large text-to-image generative models~\cite{ramesh2022hierarchical, saharia2022photorealistic,rombach2021highresolution, Yu2022ScalingAM} have achieved exciting breakthroughs, demonstrating an unprecedented ability to generate plausible images that are semantically aligned with given text prompts.
Trained on extensive datasets, these models are thought to possess a basic understanding of the world~\cite{adams2022survey}. Therefore, we aim to harness the power of these large generative models to provide unambiguous visual goals for robotic tasks without any in-domain training.

In the past, a common approach to leverage off-the-shelf large generative models in robot learning involves using these models for data augmentation on expert datasets to improve policy generalization~\cite{mandi2022cacti, chen2023genaug, yu2023scaling}. Despite the success, these methods still rely on human demonstrations, not generative models, as the main source of guidance. 

DALL-E-Bot~\cite{kapelyukh2022dall}, an early attempt to utilize web-scale generative models in a zero-shot manner for guiding manipulation tasks, utilizes DALL-E 2~\cite{ramesh2022hierarchical} to generate goal images for object rearrangement tasks. However, the generated images are often too diverse and largely disagree with real-world scenarios, requiring segmentation masks for object matching and a rule-based transformation planner to close the visual discrepancy.

The recent uprise of language-based image editing methods~\cite{hertz2022prompt, kawar2022imagic, ma2023directed} provides a new paradigm: editing the generated images to align with language descriptions, such as ``wiping'' a stain off a table, while keeping the visual appearance of the other objects and the background roughly unchanged. However, the edited images are usually examined by visual appearance rather than embodied tasks.

In this work, we introduce Learning from the Void~(\ours), a method that uses image editing techniques on pre-trained diffusion models to generate visual goals for reinforcement learning. \ours\ contains a unique image editing module capable of performing appearance-based and structure-based editing to generate goal images according to language instructions. With these goal images, \ours improves upon example-based RL methods~\cite{fu2018variational,singh2019end} to solve several robot control tasks without the need for any reward function or demonstrations.

We evaluate \ours on three simulated environment tasks and the corresponding real-world scenarios. Empirical results show that \ours can generate goal images with higher fidelity to both the editing instructions and the source images when compared to existing image editing techniques, leading to better downstream control performance in comparison to language-guided and other image-editing-based methods. We also validate the feasibility of \ours\ in real-world settings: \ours\ can retrieve meaningful reward signals from generated images comparable to those from true goal images. Based on these observations, we discuss key considerations for the current generative model research when aiming for real-world robotic applications. 

Our contributions are threefold: First, we propose an effective approach to leverage the knowledge encapsulated in large pre-trained generative models and apply it in a zero-shot manner to guide robot learning tasks. Second, we provide empirical evidence that suggests \ours -edited images provide guidance more effectively than other methods including the direct usage of text prompts. Lastly, we identify the existing gap between the capabilities of current text-conditioned image generation models and the demands in robotic applications, thus shedding light on a potential direction for future research in this domain.

%% file: 02_related_works.tex
\section{Related work}
\paragraph{Image editing for diffusion models.} Recent work in text-to-image diffusion models has demonstrated a strong ability to generate images that are semantically aligned with given text prompts~\cite{ho2020denoising, song2020denoising, ramesh2022hierarchical, saharia2022photorealistic, rombach2021highresolution, dhariwal2021diffusion, nichol2021glide}. Based on these generative models, Prompt-to-Prompt proposes to perform text-conditioned editing through injection of the cross-attention maps. Imagic \cite{kawar2022imagic} proposes to perform editing by using interpolation between source and target embeddings to generate images with edited effects. InstructPix2Pix~\cite{brooks2022instructpix2pix} demonstrates another approach for training a diffusion model on the paired source and target images and the corresponding editing instructions. Apart from editing techniques, several methods focusing on controlled editing have been proposed. Textual-Inversion~\cite{mahmoudieh22a} and DreamBooth~\cite{ruiz2022dreambooth} both aim to learn a special token corresponding to a user-specified object and at inference time can generate images containing that object with a diverse background. Directed Diffusion~\cite{ma2023directed} focuses on object placement generation, which can control the location of a specified object to reside in a given area.

\paragraph{Example-based visual RL.}
Standard reinforcement learning methods require a predefined reward function to guide the agent towards desired goal states, yet these reward functions may not always capture the essence of the task at hand or be readily available. Example-based RL methods aim to utilize the observations of the goal states to guide the learning process. VICE, proposed by Fu et al.~\cite{fu2018variational}, establishes a general framework for learning from goal observations only: it trains a discriminator with the observations in the replay buffer as negative samples and goal images as positive samples. The positive logits for new observations are used as a reward to guide the agent toward the goal observation. Building upon this, Singh et al.~\cite{singh2019end} introduce VICE-RAQ, integrating the VICE algorithm with active querying methods through periodically asking a human to provide labels for ambiguous observations. A label smoothing technique is also proposed to enhance reward shaping. Other works~\cite{li2021mural, cho2023outcome} are proposed to improve the reward shaping of VICE as well. In a parallel direction, Eysenbach et al. \cite{eysenbach2021replacing} develop Recursive Classiﬁcation of Examples (RCE), a method that learns the Q function of actor-critic methods directly from the goal observations. RCE derives the Bellman equation for discriminator-based rewards and makes temporal difference updates on them to get an estimation of the discriminator-deduced Q function directly.

\paragraph{Large generative models for robot control.} 
Current application of large generative models in robotic tasks mainly focuses on using Large Language Models for planning~\cite{ahn2022can, brohan2022rt, attarian2022see, huang2022inner, ding2023task} or learning a language conditioned policy~\cite{shridhar2022cliport, jiang2022vima}, while the application of pre-trained image generative models for robotics tasks are limited. A common way of leveraging large-scale diffusion models for robot learning is by using diffusion models to perform data augmentation on training data, as suggested in CACTI \cite{mandi2022cacti}, GenAug \cite{chen2023genaug} and ROSIE \cite{yu2023scaling}. Another line of work suggests using diffusion models to generate plans to solve robotics tasks~\cite{liu2023structdiffusion, gkanatsios2023energy, mishra2023reorientdiff}. More recently, UniPi\cite{dai2023learning} trains a text-to-video generation model on web-scale datasets and expert demonstrations to generate image sequences for planning and inverse modeling. The most related work is DALL-E-Bot \cite{kapelyukh2022dall}, which uses the DALL-E 2\cite{ramesh2022hierarchical} model to generate goal plans and performs object rearrangement according to these goals. However, unlike our work, DALL-E-Bot does not apply image editing techniques for goal generation and requires rule-based matching and predefined pick-and-place actions to bridge the visual gap between generated images and true observations.

%% file: 03_methods.tex
\section{Method}

In this study, we aim to provide zero-shot visual goals to reinforcement learning agents for manipulation tasks using only text prompts. Our approach utilizes the information grounded in large-scale pre-trained visual generative models to create semantically meaningful visual goals. We first generate a synthetic goal image dataset from raw observations using image editing techniques, then employ example-based visual reinforcement learning that is optimized for our task with the generated dataset. 

\begin{figure}[t]
    \centering
    \includegraphics[width=1.0\textwidth]{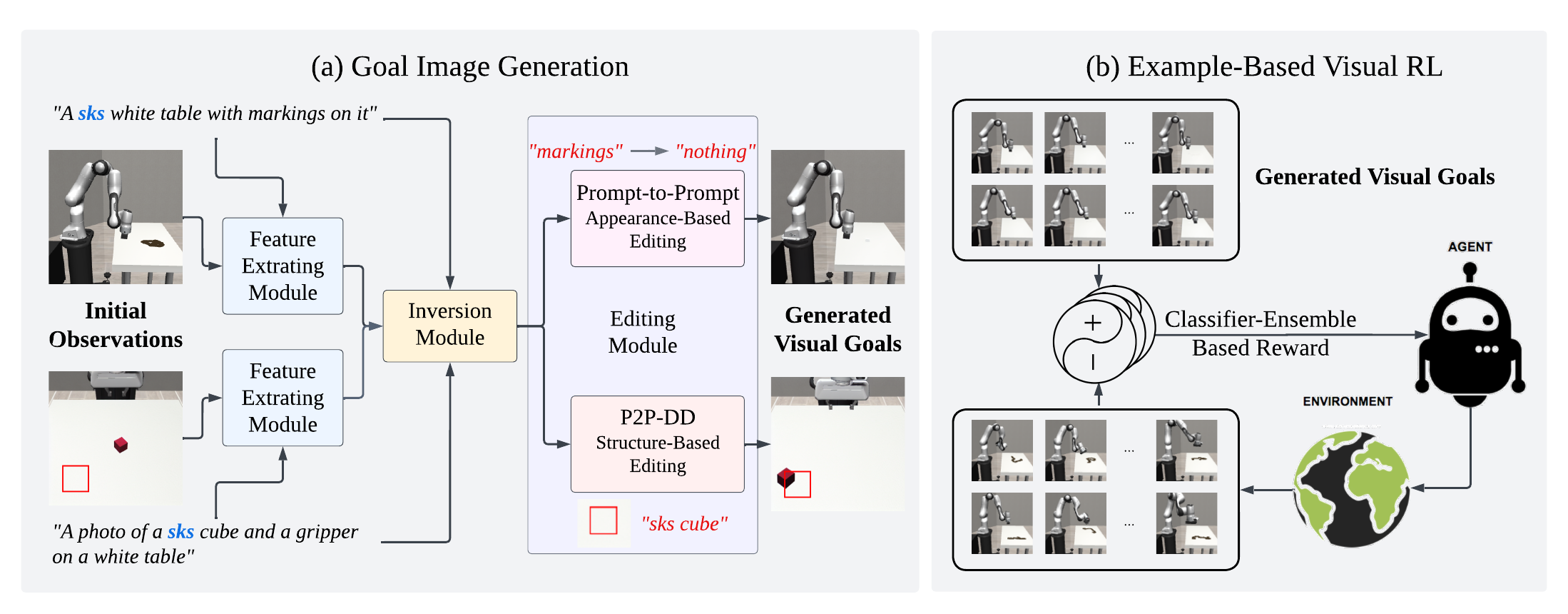}
    \vspace{-0.2in}
    \caption{\textbf{An overview of \ours.} \ours\ consists of two parts: (a) Goal image generation, where we apply image editing on the initial observations according to different editing instructions to obtain a visual goal dataset; (b) Example-Based Visual RL, where we perform reinforcement learning on the generated dataset to achieve the desired goal image in various environments.}
    \label{fig:overview}
\end{figure}

The structure of this section unfolds as follows: Section~\ref{subsec:visual-goal-generation} elucidates the image editing techniques and adjustments applied to create goal images from text prompts and initial image observations, as depicted in Figure~\ref{fig:overview}(a). Section~\ref{subsec:example-based-visual-rl} discusses the execution of example-based visual reinforcement learning using the generated goal images, as outlined in Figure~\ref{fig:overview}(b).

\subsection{Visual goal generation}
\label{subsec:visual-goal-generation}

In the first phase of our methodology, we edit and generate goal images from raw observations based on natural language guidance. Given a source prompt $\mathcal{P}$, a source image ${x_{src}}$, and an editing instruction, \ours\ synthesizes a target image ${x_{tgt}}$ using a pre-trained Latent Diffusion Model  (LDM)~\cite{rombach2021highresolution}. We consider two types of editing instructions: either a target prompt $\mathcal{P^*}$ describing the appearance changes in the target image, or a bounding-box region $\mathcal{B}$ and a set of tokens $\mathcal{I}$ corresponding to the object to be relocated when structural changes are needed.

To provide sufficient visual guidance for downstream reinforcement learning tasks, the generated goal images should highlight the visual changes while preserving the irrelevant scene as much as possible, which is a challenging requirement even for state-of-the-art image editing techniques. To this end, we integrate a number of different techniques in the visual goal generation pipeline of \ours, consisting of a feature extracting module, an inversion module, and an editing module, as outlined in Figure~\ref{fig:overview}(a). Following conventional notations, we denote $x_t$ as the synthesized LDM image at time step $t \in \{T, T-1, ..., 0\}$, where $T$ is the total diffusion time steps. Specifically, $x_T$ denotes the Gaussian noise, and $x_0$ denotes the generated image. The techniques used in each module and their purpose are described as follows.

\subsubsection{Feature extracting module}
To ensure high fidelity to the source image $x_{src}$, we learn a unique token $sks$ that encapsulates the visual features of objects within $x_{src}$ as in DreamBooth~\cite{ruiz2022dreambooth}. This process involves optimizing the diffusion model parameters along with the special token $sks$, using a set of images that contain the target object. As a result, we are able to derive a specialized model that can accurately retain key details of $x_{src}$, such as the color and texture of a cube to be positioned, or the shape of a Franka robot arm for manipulation. As later demonstrated in our experiments, this module substantially boosts the resemblance of our edited images to the corresponding source images. 

\subsubsection{Inversion module}
After employing the special token $sks$ to capture the essential features of the source scene, we invert the provided source image $x_{src}$ to a diffusion process. A simple inversion technique using the Denoising Diffusion Implicit Model (DDIM)~\cite{song2020denoising} sampling scheme calculates the samples $x_T, x_{T-1}, ..., x_0$ by reversing the ODE process. However, the inverted image $x_0$ obtained this way often deviates from $x_{src}$ due to cumulative errors.
Null-text inversion~\cite{mokady2022null} aims to mitigate this discrepancy by fine-tuning the "null-text" embedding used in 
the classifier-free guidance~\cite{ho2022classifier} $\Phi_t$ for each $t \in \{T, T-1, ..., 1\}$ to control the generation process. Thus, with the initial noise sample $x_T$ obtained through DDIM inversion and the optimized "null-text" embedding $\Phi_t, \Phi_{t-1}, ..., \Phi_1$, the diffusion process is able to generate the image $x_{src}'$, an approximation of the source image $x_{src}$.
Based on the diffusion process obtained with the inversion module, \ours can perform precise and detailed editing control on the source image, which will be discussed in Section~\ref{subsubsec:editing-module}.

\subsubsection{Editing module}
\label{subsubsec:editing-module}

Depending on the specific requirements, we divide the editing tasks into two distinct categories: appearance-based and structure-based image editing. Appearance-based editing involves maintaining the structural layout of the image while altering certain visual aspects, such as cleaning the surface of a table or lighting up an LED bulb. In contrast, structure-based editing involves changes in the layout of the image, such as relocating objects within the image from one area to another.
 
\paragraph{Appearance-based editing.} In tasks involving appearance-based editing, we employ the Prompt-to-Prompt~ editing control technique. When the Latent Diffusion Model (LDM) generates an image $x_0$ conditioned on a text prompt $\mathcal{P}$, the information encapsulated in $\mathcal{P}$ influences the diffusion process via the cross-attention layers~\cite{rombach2021highresolution, saharia2022photorealistic, hertz2022prompt}. Prompt-to-Prompt reveals that the spatial configuration of the image $x_0$ largely depends on the cross-attention maps ${M_t}$ within these cross-attention layers, especially during the initial diffusion steps.
Consequently, it suggests replacing the attention maps $M^*_t$ of the target image diffusion process with the attention maps $M_t$ from the source image diffusion process for the first $N$ time steps, beginning at time step $t=T$:

\begin{equation}
\mathrm{P2PEdit}(M, M^*, t) = \left\{
\begin{array}{lll}
M_t  &   &  {\mathrm{if} \  t \textgreater T-N} \\
M^*_t  &   &  {\mathrm{otherwise}}
\end{array} \right.
\label{equation: ptp-edit}
\end{equation}

This method ensures that the structure of the source image encapsulated in the attention maps $M_t$ is conserved in the target image in a more controlled manner, which is crucial in downstream RL tasks. 

\paragraph{Structure-based editing.} 
A notable limitation of Prompt-to-Prompt editing is the incapacity to spatially move existing objects across the image, i.e., implement structural changes. Therefore, we introduce a novel approach termed P2P-DD for structure-based editing tasks. P2P-DD combines Prompt-to-Prompt control with the idea of Directed Diffusion~\cite{ma2023directed}, extending its capability to perform object replacement.

Directed Diffusion~\cite{ma2023directed} illustrates the possibility of achieving object replacement through direct editing of attention maps during the first $N$ time steps of the generation process. The method performs attention strengthening on the cross-attention maps corresponding to the tokens in $\mathcal{I}$ (recall that $\mathcal{I}$ is the token sets representing the object of interest), through calculating a Gaussian strengthening mask (SM) of the bounding-box region $\mathcal{B}$. It also performs attention annealing to the remaining area $\mathcal{\bar{B}}$ by applying a constant weakening mask ($\text{WM}$), and the two masks are weighted by a scalar $c$:
\begin{equation}
\mathrm{DDEdit}({M_t}, \mathcal{B}, \mathcal{I}) = \left\{
    \begin{array}{lll}
    {M}_{t} \odot \text{WM}(\mathcal{\bar{B}, \mathcal{I}}) + c \cdot \text{SM}(\mathcal{B}, \mathcal{I})  &  {\mathrm{if} \  t \textgreater T-N} \\
    M_t   &  {\mathrm{otherwise}}
    \end{array} \right.
    \label{equation: dd-edit}
\end{equation}

\begin{figure}[t]
    \centering\includegraphics[width=1.0\textwidth]{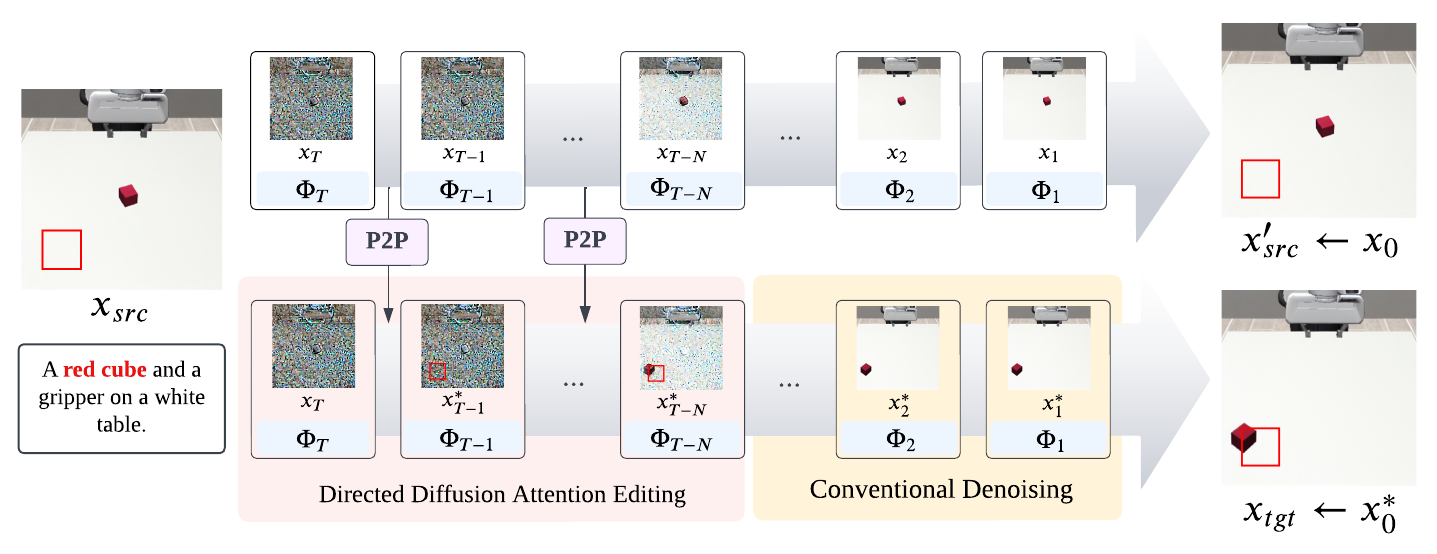}
    \vspace{-0.2in}
    \caption{\textbf{Image editing with structural changes.} The details of P2P-DD technique used in the editing module of \ours\ to perform structural changes. \ours\ performs a combination of P2P attention map injection with Directed Diffusion attention editing on the diffusion process generating the target image $x_{tgt}$, based on the Gaussian noise $x_T$ and the optimized null-text embeddings ${\Phi_T, \Phi_{T-1}, ..., \Phi_1}$ obtained through the Inversion Module of \ours.}
    \label{fig:image-editing-with-structual-changes}
\end{figure}

Our proposed P2P-DD method aims to encourage the background and other details of the image generated by Directed Diffusion with a higher resemblance to the source image. P2P-DD first injects cross-attention maps $M_t$  into $M^*_t$ as suggested by Prompt-to-Prompt. This will preserve the structure and background information from the source image, providing a decent starting point for attention map editing. Next, P2P-DD performs pixel value strengthening and weakening on the attention maps $M_t$ according to the bounding-box $\mathcal{B}$ and the tokens in $\mathcal{I}$, as suggested in the Directed Diffusion, to achieve the desired object placement:

\begin{equation}
\mathrm{P2P\mhyphen DDEdit}(M_t, M^*_t, \mathcal{B}, \mathcal{I}) = \mathrm{DDEdit}(\mathrm{P2PEdit}(M, M^*, t), \mathcal{B}, \mathcal{I})
\label{equation: ptpdd-edit}
\end{equation}

This combined editing control is only applied in the first $N$ diffusion time steps, i.e. $\{T, ..., T-N+1\}$. After the first $N$ time steps, we cease any control on attention maps and allow the target diffusion process to complete in a conventional denoising manner. The details of P2P-DD are shown in Figure ~\ref{fig:image-editing-with-structual-changes}. Attention visualization of the generation process can be found in Appendix~\ref{app: attention visualization}, and we provide a detailed description of the full algorithm in Appendix~\ref{app: full algorithm}.

\subsection{Example-based visual reinforcement learning}
\label{subsec:example-based-visual-rl}
We modify and extend the approach of VICE~\cite{fu2018variational} to devise our method for example-based visual reinforcement learning. At the core of our process, we employ the image editing methods described in Section~\ref{subsec:visual-goal-generation} on observations gathered from randomly initialized environments, producing a dataset of 1024 target images for each task. During training, these target images serve as positive samples, while the images sampled from the agent's replay buffer after a number of random exploration steps are treated as negative samples. We train a discriminator using these instances with the binary cross-entropy loss and use the output of the discriminator for new observations for the positive class as the agent reward.

The discriminator shares the CNN encoder with the reinforcement learning agent and performs classification over the output latent representations.
To improve the reward shaping, we utilize the label mixup method~\cite{singh2019end}, which performs a random linear interpolation of the 0-1 labels and their corresponding hidden vectors to obtain continuous labels between 0 and 1.
We discover that restricting the negative instances from the recent portion of the replay buffer (the last 5\%) promotes the discriminator's ability to discern subtle differences between the target and current observations. Additionally, we find that our method enjoys an ensemble of discriminators for classification results (i.e., RL rewards), which gives the agent more representative rewards.

For the reinforcement learning backbone algorithm, we use DrQ-v2~\cite{yarats2021image} for visual RL training. Based on Twin Delayed DDPG (TD3)~\cite{Fujimoto2018AddressingFA}, DrQ-v2 enhances image representation by applying random crop augmentation to the input images. Collectively, these refinements constitute our approach to example-based visual reinforcement learning, making the ``learning'' from an initial observation and language prompts possible.

%% file: 04_experiments.tex
\section{Experiments}
\begin{figure}[t]
    \centering
    \includegraphics[width=1.0\textwidth]{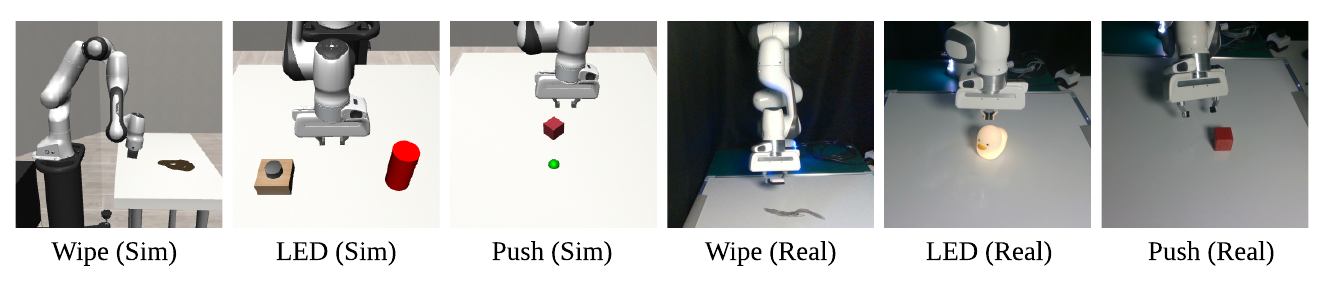}
    \caption{\textbf{Visualization of robot manipulation tasks.} We evaluate \ours on three simulation tasks: Wipe, Push, and LED, as well as three real-robot environments correspondingly.}
    \label{fig:tasks}
\end{figure}

In this section, we present a comprehensive evaluation of \ours across both simulated environments and real-world robotic tasks. An illustration of each task can be found in Figure~\ref{fig:tasks}.

The environments we use are: 1) ~LED-light, where the robot reaches for a switch or touches the light directly to turn the light from red to green; 2) ~Wipe, where the robot needs to wipe out stains from the table; 3) ~Push, where the robot needs to push a red cube to a goal position indicated by a green dot. The simulated tasks are developed based on the Robosuite benchmark, while we provide corresponding real-world tasks for each environment. A full description of the environments is provided in Appendix~\ref{app: environment_settings}.

\subsection{Goal generation}
In this section, we evaluate the ability of \ours\ and other baseline methods to generate goal images according to two types of editing instructions: appearance-based editing and structure-based editing.

\begin{figure}[t]
    \centering
    \includegraphics[width=1.0\textwidth]{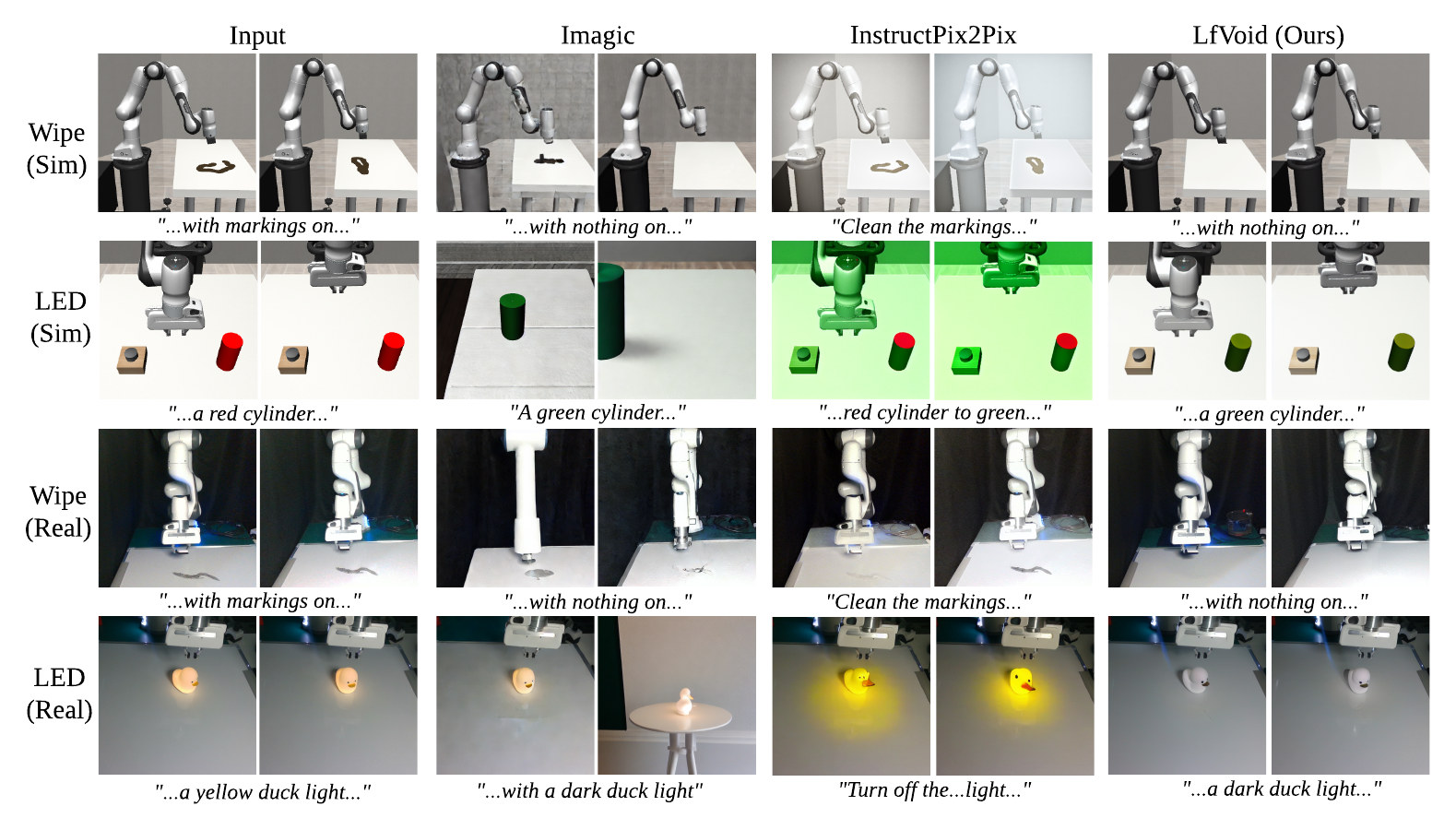}
    \vspace{-0.2in}
    \caption{\textbf{Appearance-based goal image editing results.} We compare \ours against two recent image editing methods, Imagic and InstructPix2Pix, on four editing tasks. In both simulation and real settings, \ours can generate images that are better aligned with given text prompts while preserving the remaining source scene. See Appendix~\ref{app: additional_results} for more examples.}
    \label{fig:goal_generation_1}
\end{figure}

\paragraph{Appearance-based goal image editing results.} In Figure~\ref{fig:goal_generation_1}, we present the visual goal generation results of the "Wipe" and "LED" tasks, focusing on editing object appearances in both simulated and real-world environments. We compare our method with two existing image editing methods: Imagic~\cite{kawar2022imagic} and InstructPix2Pix ~\cite{brooks2022instructpix2pix}. Imagic struggles to preserve the texture and details of the image background, often resulting in significant alterations to the source image. InstructPix2Pix fails to effectively locate the specified regions indicated by the editing instructions, frequently causing global color changes and failing to perform the proper edits. Our method demonstrates superior performance in terms of maintaining the appearance of unrelated parts while effectively editing specified regions. As mentioned in Section~\ref{subsubsec:editing-module}, through directly applying control and editing on the attention maps, \ours is capable of locating the target area and performing the desired edit based on language instructions. 

\begin{figure}[!t]
    \centering
    \includegraphics[width=1.0\textwidth]{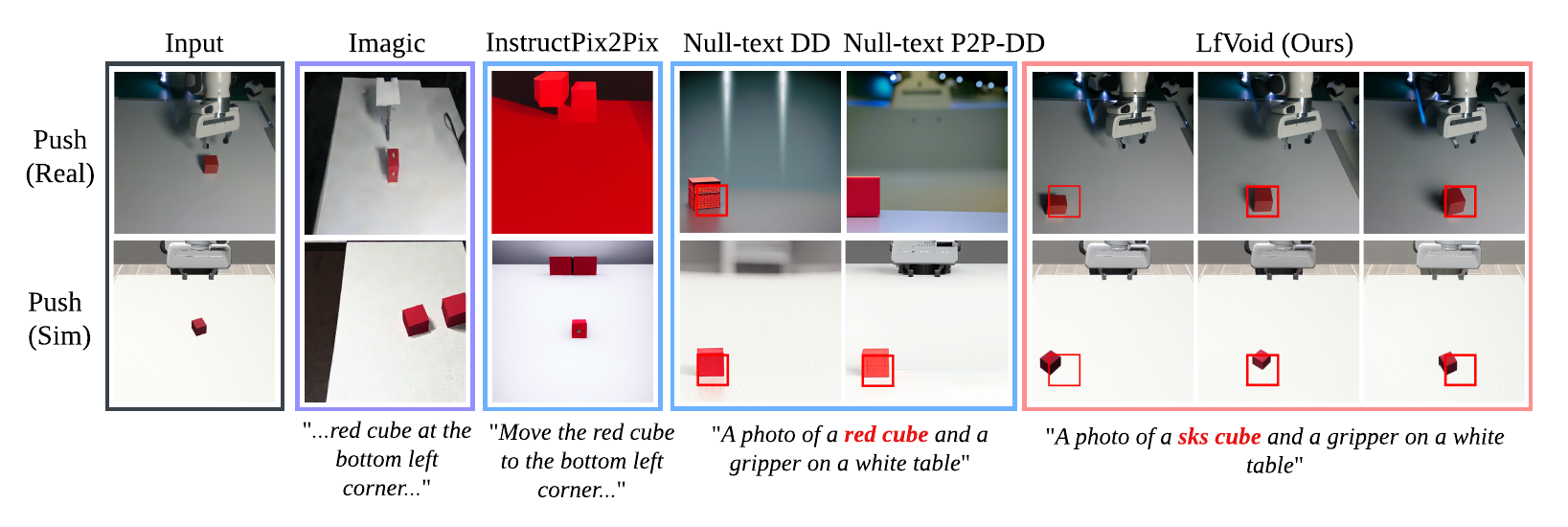}
    \vspace{-0.2in}
    \caption{\textbf{Structure-based goal image editing results.} For structure-based editing tasks, we compare \ours against Imagic and InstructPix2Pix, as well as two ablations, Null-text Directed Diffusion~(Null-text DD) and Null-text P2P-DD. \ours can successfully perform object displacement and preserves the background and details of the source image, while other methods fail to do so. See Appendix~\ref{app: additional_results} for more examples.}
    \label{fig:goal_generation_2}
\end{figure}

\paragraph{Structure-based goal image editing results.} Next, we report the performance of \ours in editing tasks with structural changes ("Push") in both simulated and real-world settings.  In Figure~\ref{fig:goal_generation_2}, we first present the editing results of Imagic and InstructPix2Pix: Imagic distorts the robot arm and table significantly and often introduces multiple objects into the image, rather than moving the existing one. InstructPix2Pix only changes the overall color of the image or replaces the gripper of the robot arm with a geometric body, failing to achieve the required object displacement. In addition, we also present ablative results of \ours when certain modules are removed: Null-text DD removes the feature extracting module and P2P in the editing module. Although it can successfully move objects to the lower-left corner, the background and the robot arm nearly disappear. Null-text P2P-DD removes only the feature extracting module of \ours, and results show that it improves the preservation of the source image's background compared to Null-text DD, but the shape of the moved object is inconsistent with its original form. Our method, when compared to all baselines, can perform the object displacement task successfully while better preserving the visual features of the object and the remaining scene. Moreover, we also demonstrate \ours's ability to perform object displacement at different locations, as shown in the last three columns of Figure~\ref{fig:goal_generation_2}. These experiments reveal the importance of each component in our method. The creative integration of different image editing techniques provides a powerful tool for image editing with structural changes, demonstrating a significant improvement over existing methods.

\paragraph{Human user evaluation.} In Table~\ref{table: user study}, we present the user study results in terms of normalized Elo score. The evaluators show a strong preference for images generated by \ours compared to other methods. Details of the user study and the Elo metrics can be found in Appendix~\ref{app: user study}.
\vspace{-10pt}
\begin{table}[!hb]
  \centering
  \begin{tabularx}{\textwidth}{cccccc}
    \toprule              
    & Imagic     & InstructPix2Pix    & NT-DD & NT-P2P-DD & Ours \\
    \midrule
    Appearance-Based ($\uparrow$) & 47.7  & 42.4 & N/A & N/A & \textbf{87}    \\
    Structure-Based ($\uparrow$) & 43.6 & 18.2 & 58 &  63.8 &  \textbf{91.6} \\
    \bottomrule
  \end{tabularx}
  \caption{\textbf{User study on visual goal generation.} The Elo score shows user preference among multiple choices, higher is better.}
  \label{table: user study}
\vspace{-0.5cm}
\end{table}

\subsection{Example-based visual reinforcement learning}
\label{subsec:example-based visual rl results}

\begin{figure}[b]

\begin{subfigure}{0.33\textwidth}
\includegraphics[totalheight=1.37in]{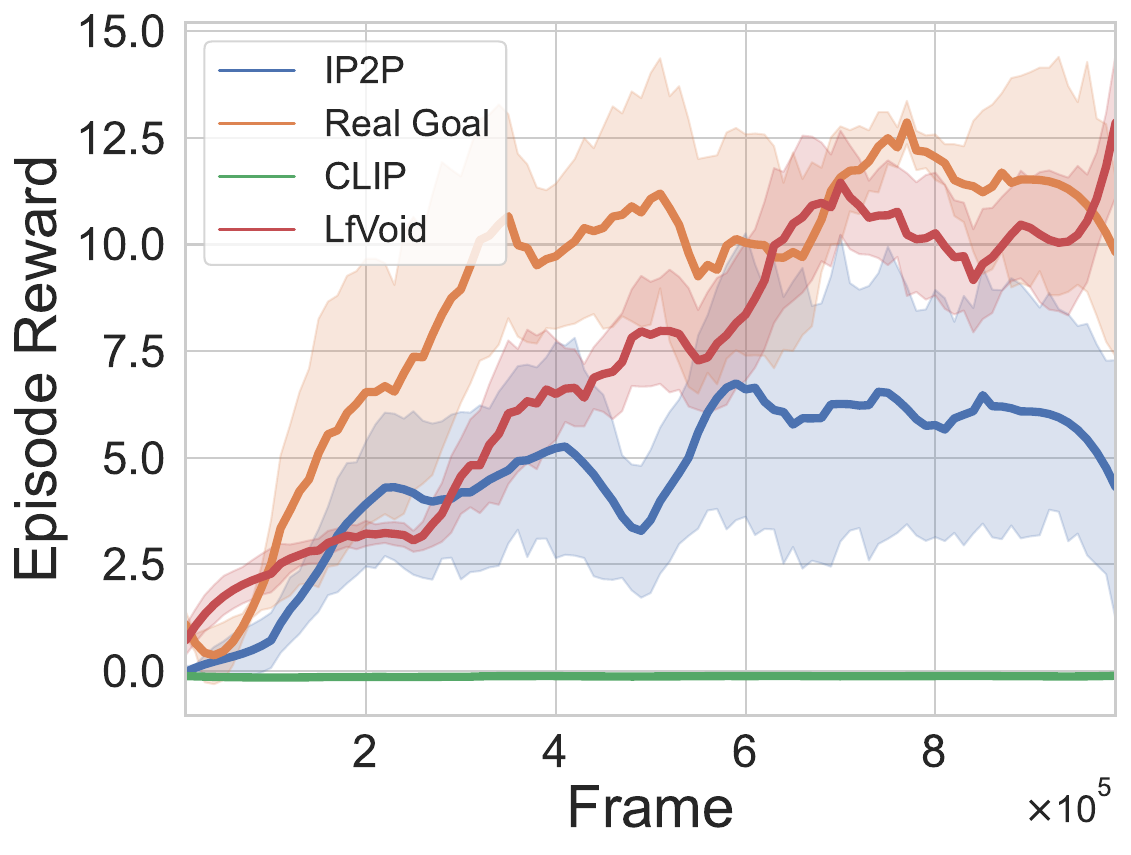} 
\caption{Wipe Episode Reward}
\label{fig:subim1}
\end{subfigure}
\begin{subfigure}{0.32\textwidth}
\includegraphics[totalheight=1.37in]{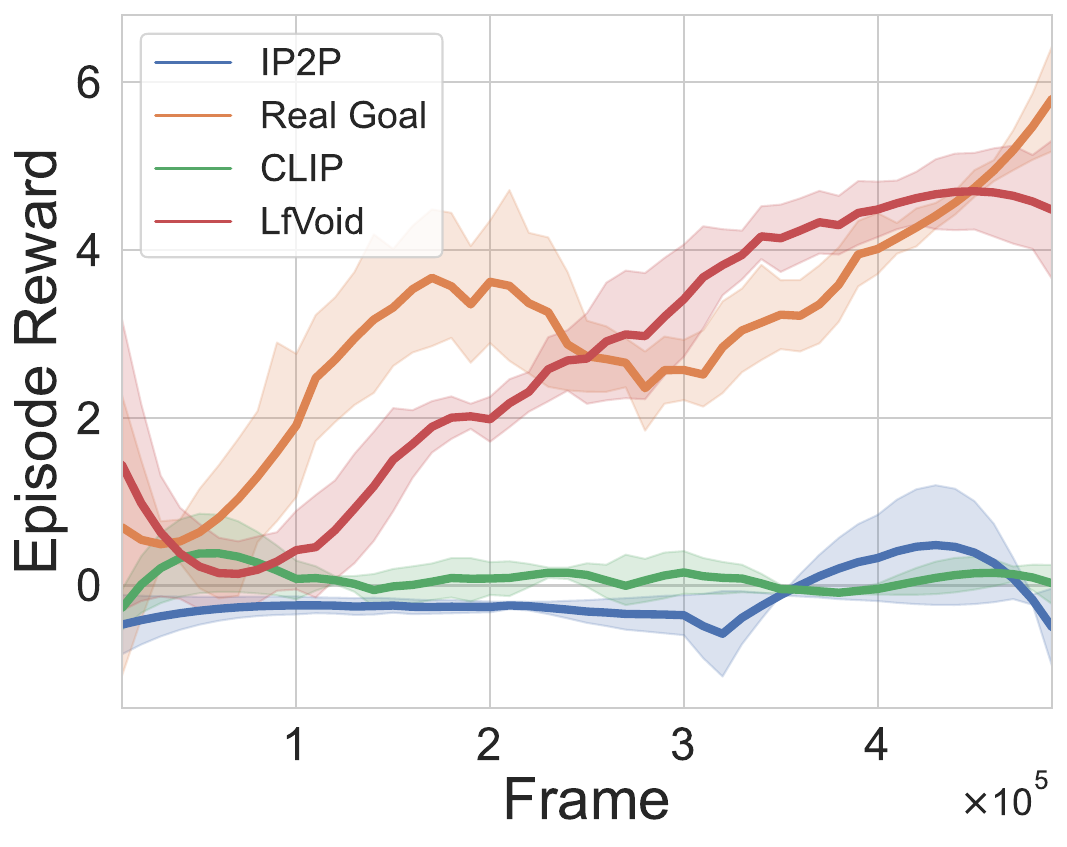}
\caption{LED Episode Reward}
\label{fig:subim2}
\end{subfigure}
\begin{subfigure}{0.33\textwidth}
\includegraphics[totalheight=1.37in]{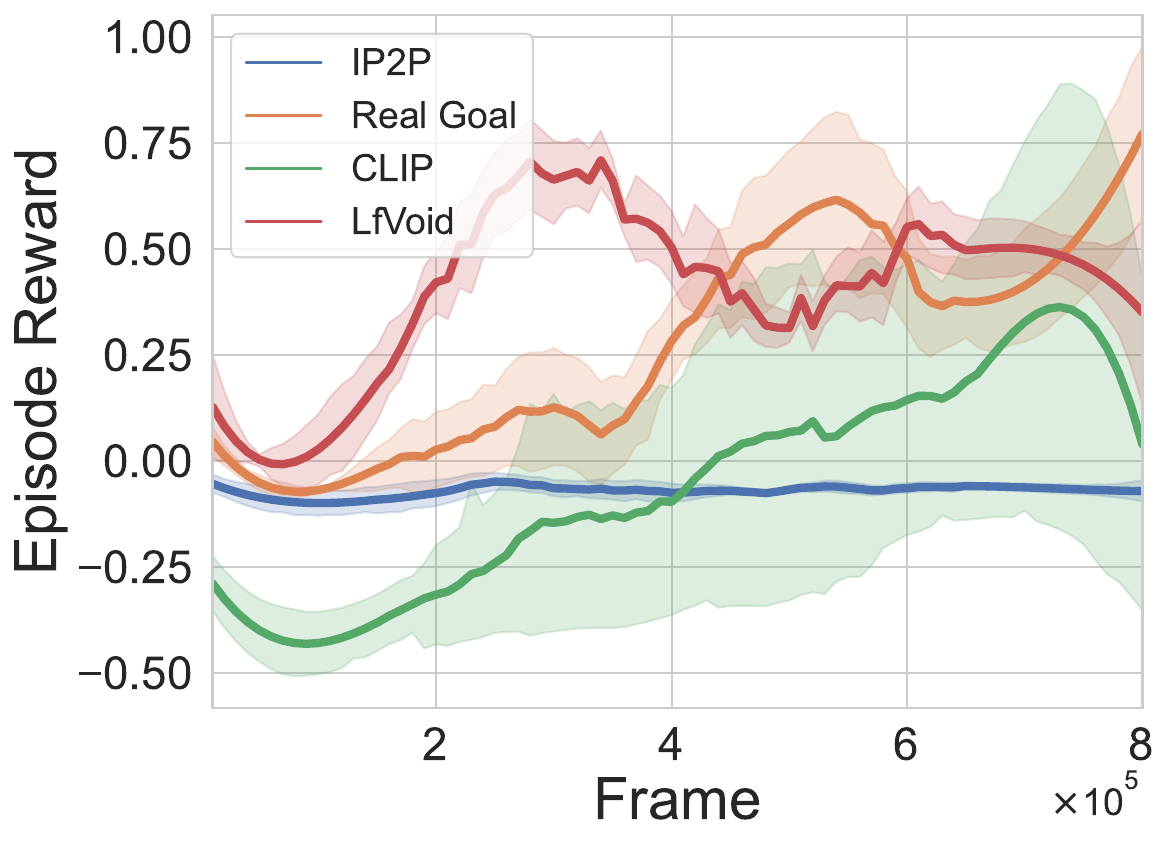} 
\caption{Push Episode Reward}
\label{fig:subim3}
\end{subfigure}
\caption{\textbf{Episode reward curve of simulation tasks.} The results of the CLIP baseline~(CLIP), InstructPix2Pix baseline~(IP2P), using real goal image as an upper bound~(Real Goal), and \ours~(Ours)}
\label{fig:episode_reward}

\end{figure}

We now dive into the results obtained from applying example-based visual reinforcement learning using the goal images generated by \ours. We select three baselines for our comparison. The first baseline uses language guidance by leveraging the Contrastive Language-Image Pre-training~(CLIP) score~\cite{radford2021learning} of the task prompt and image observations as a reward to train an RL agent. This baseline was chosen to verify the effectiveness of translating prompts into images against raw language hints. The second baseline uses InstructPix2Pix~(IP2P) for image editing and then applies the same example-based RL as in \ours on the resulting images. It allows us to compare the performance of \ours with a straightforward image editing technique. We also present an upper bound performance of our pipeline~(marked as Real Goal), where we manually create ideal goal images, such as erasing the stains directly in the simulator for the "Wipe" task. This strategy illustrates the potential of our algorithm and indicates the gap between the generated goals and the real ones.

\paragraph{Simulation results.} We report both episode reward (Figure~\ref{fig:episode_reward}) and numerical metrics (Table~\ref{table:success_rate} on three simulation tasks. CLIP baseline achieves low rewards in all the tasks, showing that directly using language guidance through CLIP embeddings cannot provide sufficient guidance. Moreover, \ours consistently outperforms the IP2P baseline and gains comparable performance with the real goal upper bound, providing evidence that generated goal images with high fidelity to both the original scene and the editing instructions is crucial for the deployment of example-based RL methods. 

\paragraph{Real-world results.} In this experiment, we aim to validate the feasibility of \ours in more complex real-world settings through visualization of reward functions. As in our previous experiment, we initially generate goal images using InstructPix2Pix (IP2P) and \ours, and concurrently collect a real goal dataset as an upper-bound (Real Goal). We then train discriminators on these datasets separately. 
We evaluate the discriminators on five successful trajectories for each task on a real robot and obtain the reward curve. In addition, we compare our approach with language-based guidance, which uses the distance of the CLIP embedding between the task prompt and the current observation image as rewards. 

\begin{figure}[t]

\begin{subfigure}{0.33\textwidth}
\includegraphics[totalheight=1.35in]{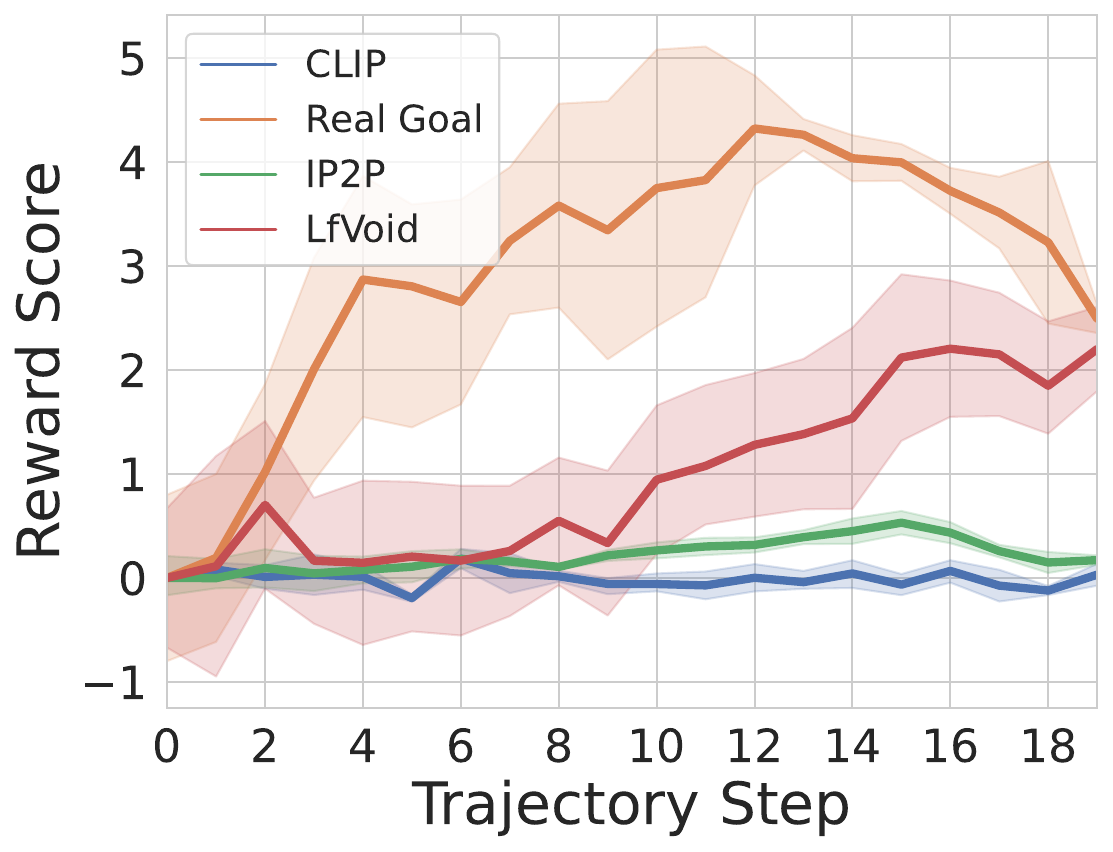} 
\caption{Wipe Reward Curve}
\label{fig:subim1}
\end{subfigure}
\begin{subfigure}{0.33\textwidth}
\includegraphics[totalheight=1.35in]{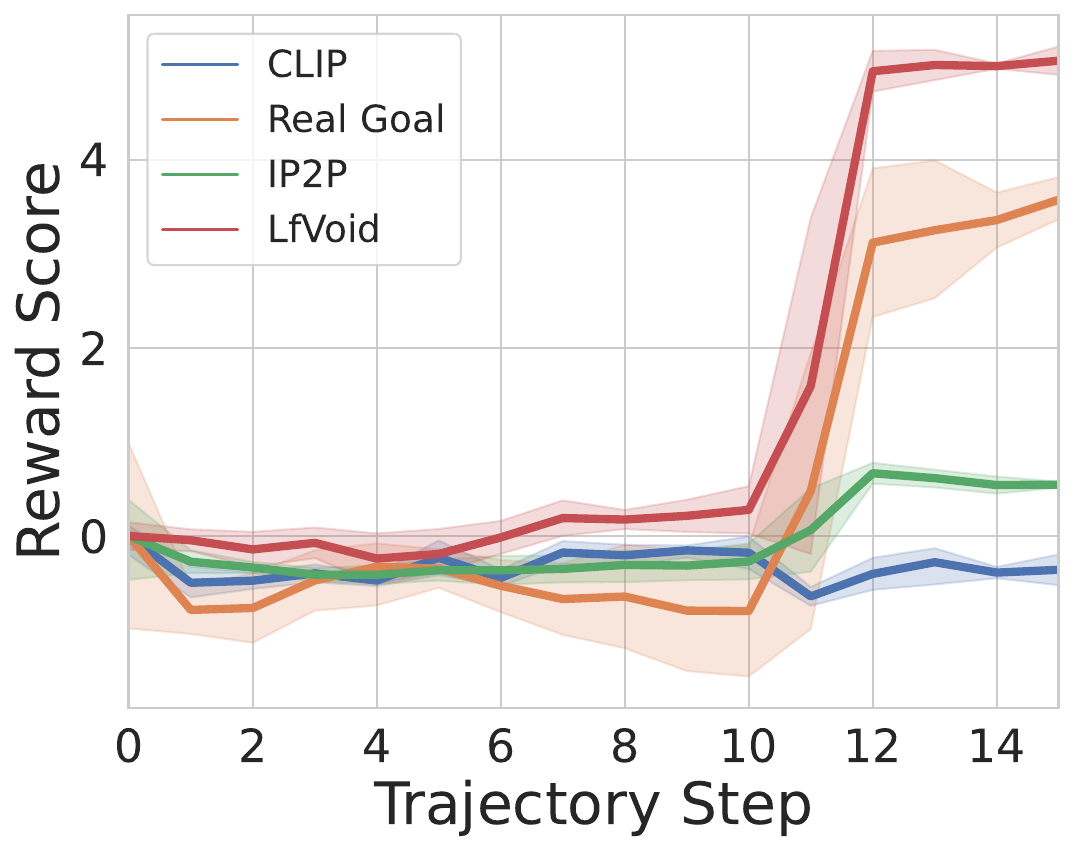}
\caption{LED Reward Curve}
\label{fig:subim2}
\end{subfigure}
\begin{subfigure}{0.33\textwidth}
\includegraphics[totalheight=1.35in]{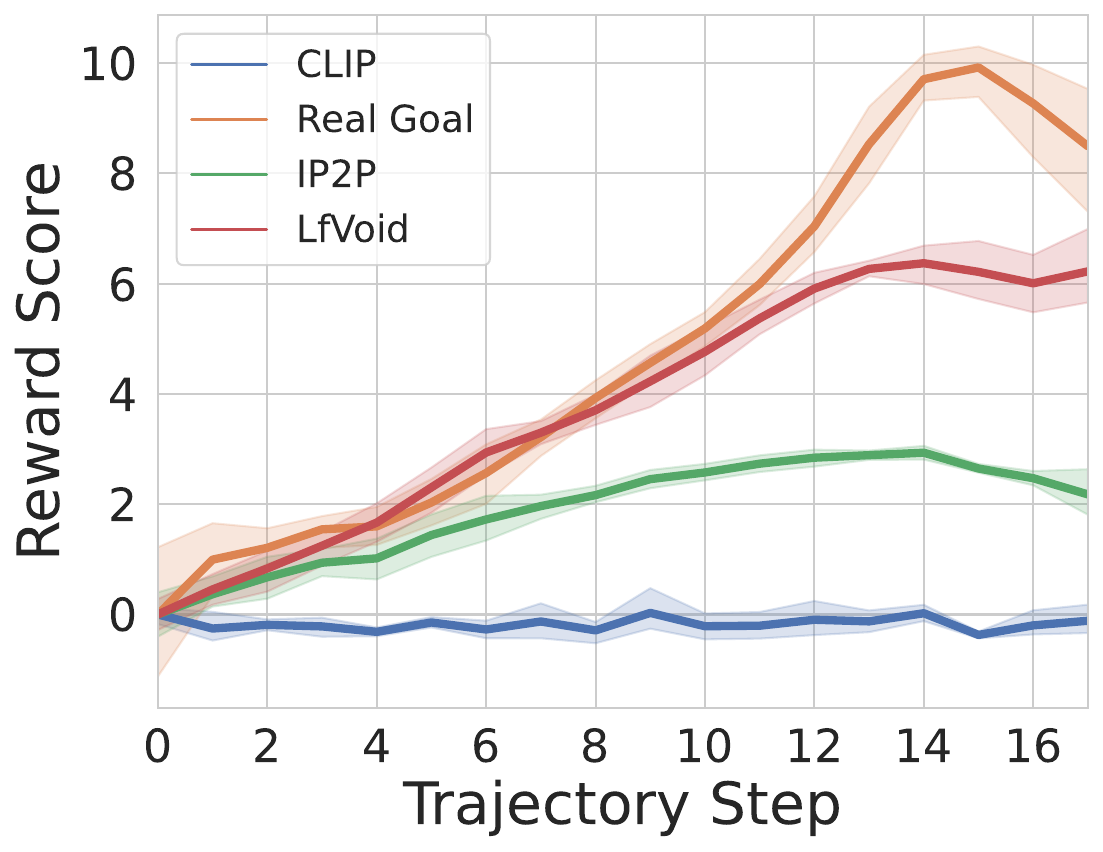} 
\caption{Push Reward Curve}
\label{fig:subim3}
\end{subfigure}

\caption{\textbf{Reward curve on successful trajectories for real-robot tasks.} Visualization of the reward function of \ours~(Ours) compared with baselines.}
\label{fig:real robot reward curves}

\end{figure}

As illustrated in Figure~\ref{fig:real robot reward curves}, the reward curve derived from \ours can provide near monotonic dense signals comparable to those from the real goals. On the contrary, the curves obtained by IP2P and the CLIP score fail to accurately track the progress of each task. This demonstrates that our method offers a more instructive learning signal for real-world robotic tasks than either IP2P or CLIP.
\begin{table}[t]
  \centering
  \begin{tabularx}{\textwidth}{ccccc}
    \toprule              
    & CLIP     & InstructPix2Pix    & Real Goal & Ours \\
    \midrule
    Wipe~(Cleaned Stains / Patch ) & 0.0±0.0 & 1.7±1.9 & 22.0±5.8 & 21.3±5.8  \\
    LED~(Success Rate / \%)  & 10.0±20.0 & 0.0±0.0 & 93.3±11.5 & 75.0±50.0   \\
    Push~(Success Rate / \%) & 12.5±19.1 & 3.3±3.5 & 30.0±13.7 & 27.9±12.7   \\
    \bottomrule
  \end{tabularx}
  \caption{\textbf{Numerical metrics of simulation tasks.} We report the success rate for LED and Push, and the number of stain patches cleaned for Wipe, please refer to Appendix~\ref{app: environment_settings} for details.}
  \label{table:success_rate}
\end{table}

%% file: 05_conclusion.tex
\section{Discussion}
In this work, we present \ours, an effective approach for leveraging the knowledge of large-scale text-to-image models and enabling zero-shot reinforcement learning from text prompts and raw observation images. We have identified and tackled numerous challenges that arise when applying state-of-the-art image editing technologies to example-based RL methods. Our work highlights the potential for the adaptation and application of image-generation techniques in the realm of robotics. Our findings not only enhance the understanding of image editing for robotic applications but also provide a clear direction for the image generation community to address real-world challenges.

Although our proposed method, \ours, has shown promising results, we acknowledge that it is not without limitations. While \ours has succeeded in generating goal images to train example-based RL policies, a certain level of prompt tuning is needed in order to achieve optimal editing performance. We refer the readers to Appendix~\ref{app: limitation} for an analysis of failure cases.

We would also like to highlight that there exists a gap between the ability of text-to-image generation models and the need for robot learning. For example, large diffusion models exhibit poor understandings of the spatial relationships between objects~\cite{conwell2022testing}; therefore, both the generative models and the editing methods struggle to handle object displacement solely through language prompts. The Directed Diffusion technique, while being able to perform movements of objects, requires a user-defined bounding box to achieve precise control. Furthermore, large generative models sometimes struggle to generate images that are considered valid under physics laws. When asked to pick up a bottle with a robot arm, the model simply stretches the shape of the arm to reach the bottle rather than changing the joint position. Lastly, AI-generated images inevitably introduce alterations to the details of objects. The robustness of current visual Reinforcement Learning (RL) algorithms to such changes remains an open question. 

%% file: 07_acknowledgement.tex
\section*{Acknowledgment}
This work is supported by  National Key R\&D Program of China (2022ZD0161700).

%% file: 06_appendix.tex
\appendix
\section{Algorithm details}
\label{app: algorithm details}

\subsection{Attention visualization}
\label{app: attention visualization}

\begin{figure}[h!]
\includegraphics[width=12cm, height=20cm]{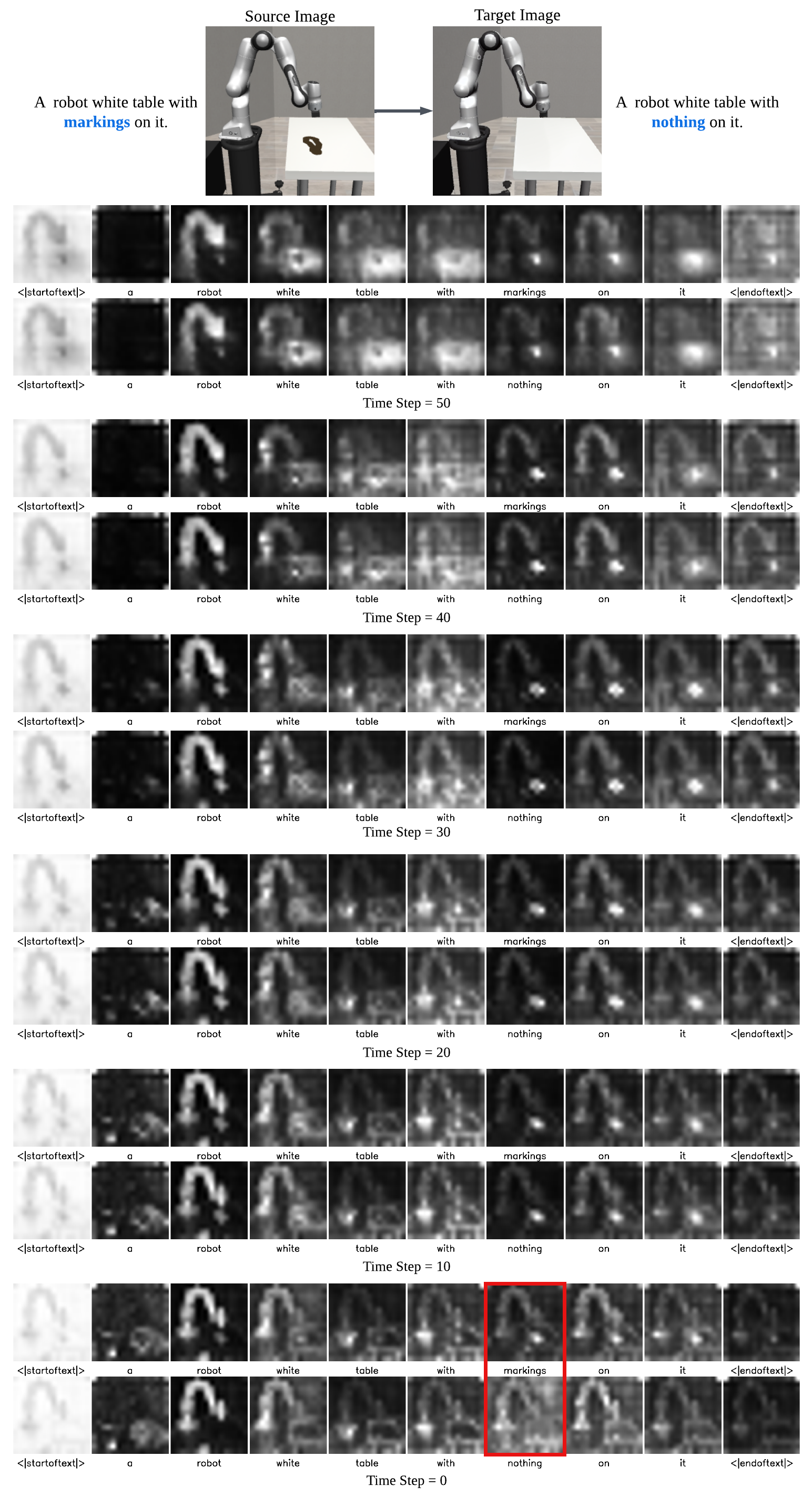}
\caption{\textbf{Attention visualization of Prompt-to-Prompt.} The visualization of cross-attention maps at diffusion step $t=50, 40, ..., 0$, with the cross-attention maps generated by the source diffusion process~(upper row) and the target diffusion process~(bottom row). The first 40 steps keep the original structure of the source image, and the required editing is performed during the last 10 diffusion steps, as indicated by the red box.}
\label{fig: attention_visualization_ptp}
\end{figure}

\paragraph{Prompt-to-Prompt editing.} In the Prompt-to-Prompt editing algorithm~\cite{hertz2022prompt}, attention editing~(P2PEdit), as defined in Equation~\ref{equation: ptp-edit}, is applied on the cross-attention maps during first $T-N$ steps, where $T=50, N=10$ for Wipe. For visualization purposes, we calculate the average cross-attention maps of size $16\times 16$ at time step $t=50, 40, 30, 20, 10, 0$ during the source image diffusion process as well as the target image diffusion process, and present the results in Figure~\ref{fig: attention_visualization_ptp}. We can observe that in time steps $t=50, 40, 30, 20, 10$, the attention maps in the upper row~(source image diffusion) are very similar to that of the bottom row~(target image diffusion): this is because the cross-attention maps from the upper row are injected to the bottom row, in order to preserve the structure of the figure. However, after the first 40 steps, we cease any control on the attention maps, therefore the attention maps of the bottom row significantly differ from the upper at time step $t=0$. This is when the editing instruction of replacing "markings" with "nothing" is performed, as indicated by the red box in Figure~\ref{fig: attention_visualization_ptp}.

\begin{figure}[!hb]
\includegraphics[width=\linewidth]{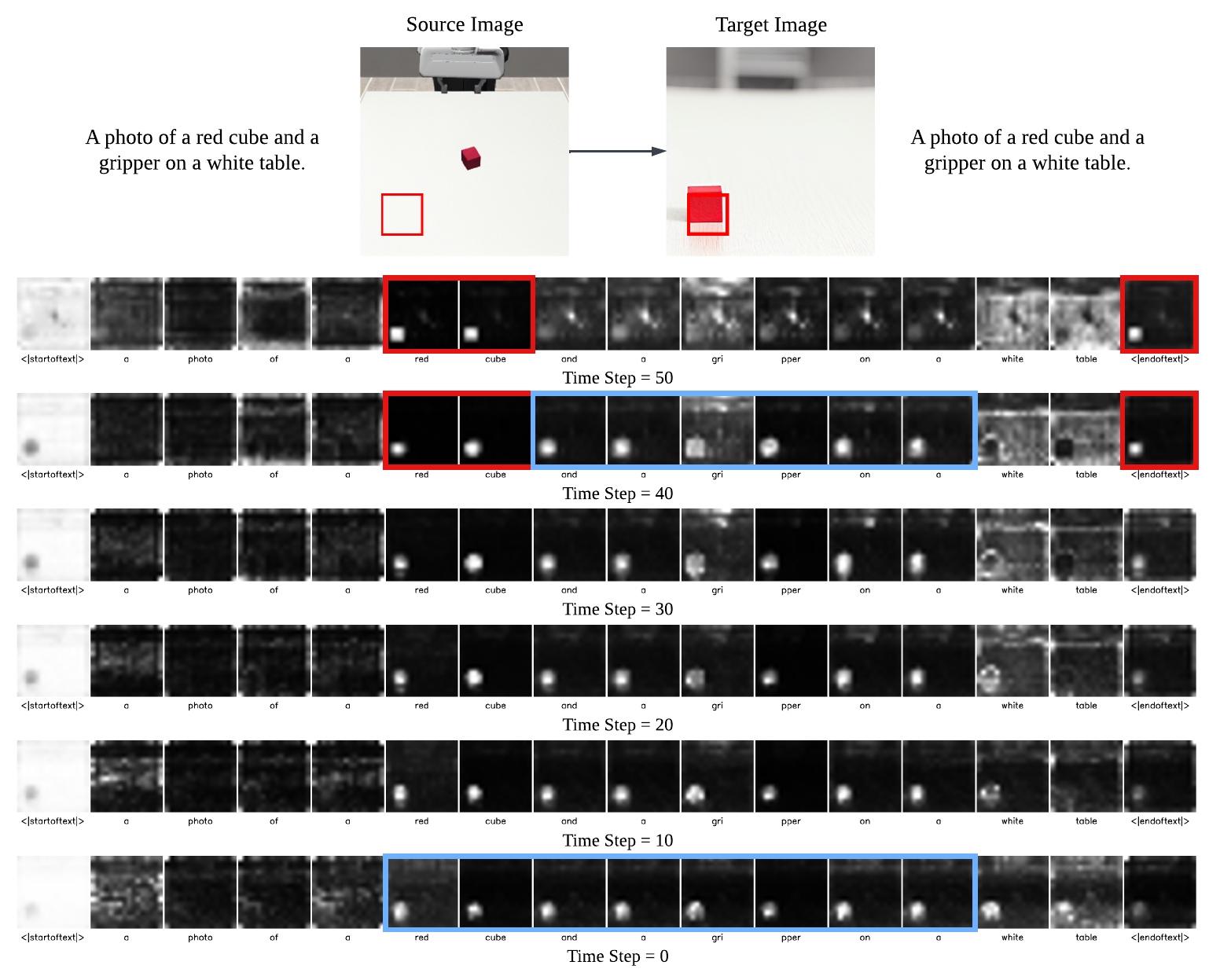}
\caption{\textbf{Attention visualization of Directed Diffusion.} The visualization of cross-attention maps at diffusion step $t=50, 40, ..., 0$. The attention strengthening and weakening suggested in Directed Diffusion are performed during the first 20 steps, as indicated by the red box. Although strengthening is initially applied on the bounding-box area, its effect will spread during the subsequent diffusion steps, leading to successful object displacement, as indicated by the blue box.}
\label{fig: attention_visualization_dd}
\end{figure}

\paragraph{Directed Diffusion editing.} In Figure~\ref{fig: attention_visualization_dd}, we present the visualization of Directed Diffusion~\cite{ma2023directed} used in the Push task. Recall that in Directed Diffusion, we perform value strengthening in the bounding-box area and weakening the other area of the attention maps during the first $T-N$ diffusion steps. For the Push task, we set $T=50, N=30$. At step $t=50, 40$, attention strengthening is performed at tokens corresponding to the word "red" and "cube", as well as the trailing tokens at the end of the sentence. After the first 20 steps, we let the diffusion process finish in a conventional denoising manner. As shown in the attention maps from time-step $t=30, 20, 10, 0$, the strengthening effect is preserved during the later diffusion steps, leading to a successful displacement. However, because we anneal the remaining area outside the bounding box, the background information of the source image is lost during the attention editing~(such as the appearance of the gripper), therefore the background of the generated target image differs largely from the source image.

\begin{figure}[t]
\includegraphics[width=\linewidth]{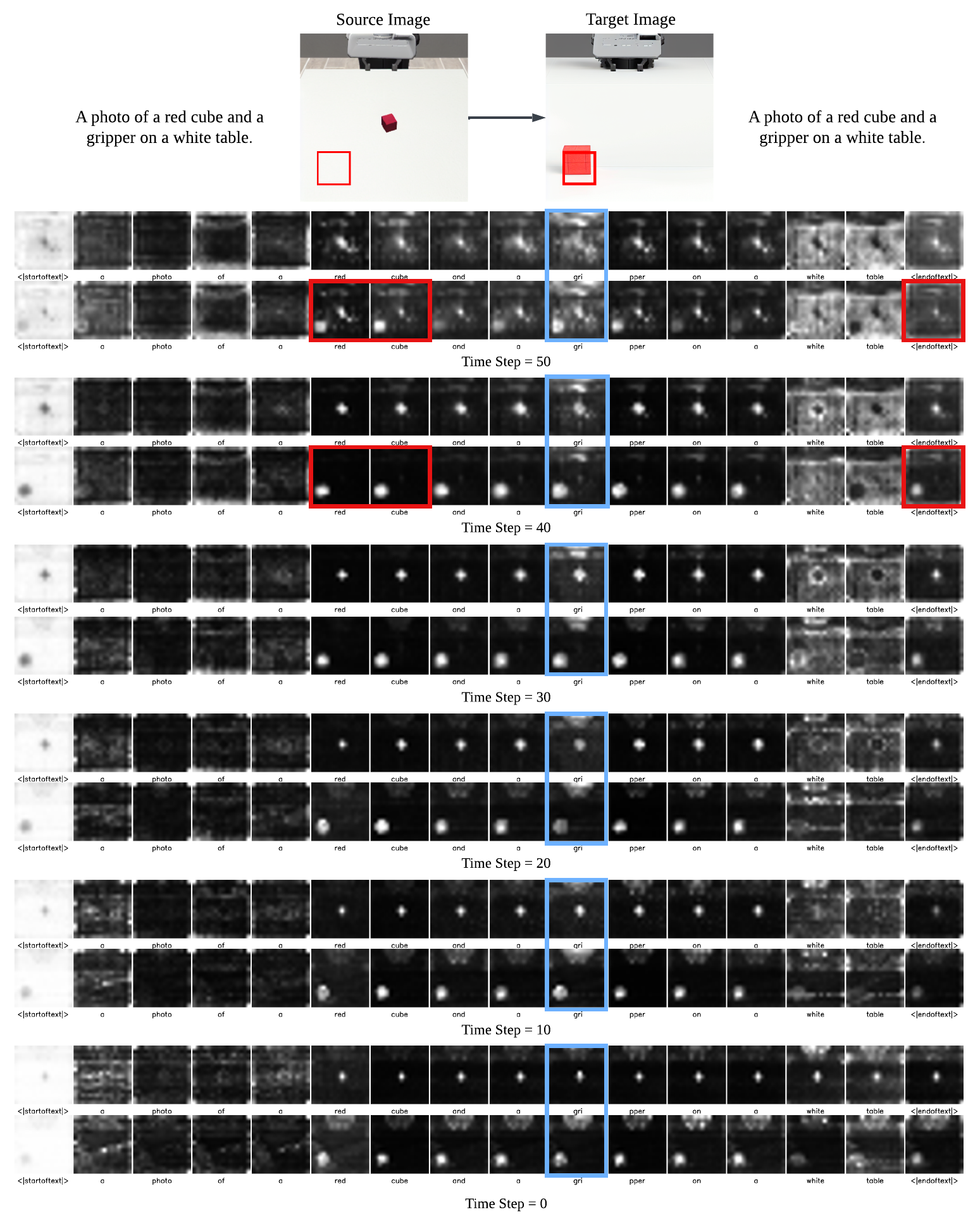}
\caption{\textbf{Attention visualization of P2P-DD.} The visualization of cross-attention maps at diffusion step $t=50, 40, ..., 0$. Attention map injection from source diffusion process~(upper row) to target diffusion process~(bottom row) is first performed at each step, preserving the background of the source image, as highlighted by the blue box. Then attention strengthening and weakening are applied on the target diffusion, as indicated by the red box, to generate an image that achieves the desired editing while preserving the original scene.}
\label{fig: attention_visualization_ptpdd}
\end{figure}

\paragraph{P2P-DD editing.} We propose the P2P-DD algorithm, a combination of Prompt-to-Prompt and Directed Diffusion techniques to achieve structure-based editing for the Push task. As defined in Equation~\ref{equation: ptpdd-edit}, P2P-DD runs two diffusion processes in parallel. As shown in Figure~\ref{fig: attention_visualization_ptpdd}, the upper row represents the attention map of the source image diffusion process, and the bottom row represents that of the target image. Instead of directly performing strengthening and weakening on the attention maps, P2P-DD first injects the attention maps from the source diffusion process and then performs editing on these injected attention maps. In this way, the details of the source image can be better preserved: the gripper in the background is preserved in the attention maps of later time steps, as indicated by the blue box in Figure~\ref{fig: attention_visualization_ptpdd}.

\subsection{Full algorithm}
\label{app: full algorithm}
The complete visual goal generation algorithm is presented in Algorithm~\ref{alg:visual-goal}, and the complete example-based visual RL algorithm is provided in Algorithm~\ref{alg:visual-rl}. 

Following conventional notations, we denote $x_t$ as the synthesized LDM image at time-step $t \in \{T, T-1, ... ,0\}$, where $T$ is the total diffusion steps. Specifically, $x_T$ denotes the Gaussian noise, and $x_0$ denotes the generated image. For notations used in Algorithm~\ref{alg:visual-goal}, $\Phi_t$ denotes the "null-text" embedding at step $t$, and $M_t$ denotes the attention maps at step $t$.

For the example-based RL part, starting from a dataset of edited target observation $\mathcal{D}^{edit}\{s_i^{edit}\}_{i=1}^{N_g}$. We  use the current policy $\pi$ to interact with the environment and collect a dataset $\mathcal{D}^{env}$, and train an ensemble of classifiers $\{\Phi^i\}_{i=1}^{N_c}$ on the linearly mixuped batch $\mathcal{B}$ consists of samples $\{s_i^{env}\}_{i=1}^{\mathcal{B}}$ from $\mathcal{D}^{env}$ as negative class, and samples from the edited images $\{s_i^{edit}\}_{i=1}^{\mathcal{B}}$ as positive class. During training, we use DrQv2 as the base RL algorithm and use the positive logits from the classifier as the reward. The update of classifiers and the RL agent take place in turn and we only use the recent $\lambda$-portion of the replay buffer (denoted as $\mathcal{D}^{env}_\lambda$) to train the classifier. The pseudo-code of example-based RL can be found at Algorithm~\ref{alg:visual-rl}. And an introduction to the mixup technique we use is provided as follows:

Consider $s_i$ and $s_j$ as any two inputs to the classifier, originating either from the replay buffer or the set of goal examples, with their associated labels $y_i$ and $y_j$. Mixup regularization employs linear interpolation to these input/output pairs to construct a synthetic training distribution as follows:
\begin{equation}
\tilde{s} = \lambda s_i + (1 - \lambda)s_j, \quad
\tilde{y} = \lambda y_i + (1 - \lambda)y_j
\end{equation}
Here, $\lambda \sim \text{Beta}(\alpha, \alpha)$ follows the Beta distribution. The mixup parameter $\alpha$ determines the degree of mixup, with higher $\alpha$ values implying a greater mixup level (i.e., the sampled $\lambda$ values are closer to 0.5 than 0). As illustrated by Zhang et.al.~\cite{zhang2017mixup}, mixup contributes to smoother transitions between distinct classes by promoting linear behavior. 

\begin{algorithm}[t]
\caption{Visual Goal Generation of \ours}
\label{alg:visual-goal}
\begin{algorithmic}[1]
\STATE \textbf{Input}: A source prompt embedding $P$, a source image $x_{src}$, a diffusion model $DM(.)$, a text encoder $\phi(.)$, a target prompt $P^*$(if appearance-based), a bounding-box region $\mathcal{B}$ and a token set $\mathcal{I}$(if structure-based)

\STATE \textbf{Output}: A target image $x_{tgt}$
\\\hrulefill

\STATE Set guidance scale $\omega=1$
\STATE Compute the DDIM inversion results $x'_T, ..., x'_0$ over the source image $x_{src}$
\STATE Set guidance scale $\omega=7.5$
\STATE Initialize $\bar x_T \leftarrow x'_T$, null-text embedding $\Phi_T \leftarrow \phi("")$
\FOR{t=T, T-1, ... 1}
    \FOR{j=0, ... N-1}
        \STATE $\Phi_t \leftarrow \Phi_t-\eta\nabla_{\Phi} || x'_{t-1} - x_{t-1}(\bar x_t, \Phi_t, \phi(P))||_2^2 $
    \ENDFOR
    \STATE Set $\bar x_{t-1} \leftarrow x_{t-1}(\bar x_t, \Phi_t, \phi(P))$, $\Phi_{t-1} \leftarrow \Phi_t$
\ENDFOR

\STATE $x_T \leftarrow x'_T$, $x^*_T \leftarrow x_T$
\FOR{$t=T, T-1, ..., 1$}
    \STATE $x_{t-1}, M_t \leftarrow \mathrm{DM}(x_t, P, t, \Phi_t)$
    \STATE $M^*_t \leftarrow \mathrm{DM}(x^*_t, P, t, \Phi_t)$
    \IF{appearance-based editing}
        \STATE $M^*_t \leftarrow \mathrm{P2PEdit}(M_t, M^*_t, t)$
        \STATE $x^*_{t-1} \leftarrow \mathrm{DM}(x^*_t, P^*, t, \Phi_t)\{M\leftarrow M^*_t\}$
    \ELSE
        \STATE $M^*_t \leftarrow \mathrm{P2P\mhyphen DDEdit}(M_t, M^*_t, \mathcal{B}, \mathcal{I})$
        \STATE $x^*_{t-1} \leftarrow \mathrm{DM}(x^*_t, P, t, \Phi_t)\{M\leftarrow M^*_t\}$
    \ENDIF
    
\ENDFOR
\STATE $x_{tgt} \leftarrow x^*_0$
\STATE Return $x_{tgt}$
\end{algorithmic}
\end{algorithm}

\begin{algorithm}[t]
\caption{Example-based Visual RL of \ours}
\label{alg:visual-rl}
\begin{algorithmic}[1]
\STATE \textbf{Input}: An edited dataset $\mathcal{D}^{edit}$, ensemble of classifiers $\{\Psi^i\}_{i=1}^{N_c}$, an DrQv2 agent $\pi$ and critic $Q_1, Q_2$, classifier training interval $t_{cls\_int}$, classifier training steps $t_{cls\_steps}$, label mixup hyperparameter $\alpha$, training environment $e$, batch size $\mathcal{B}$
\STATE \textbf{Output}: Trained DrQv2 agent $\pi$
\\\hrulefill
\STATE $s \leftarrow e.\text{reset()}$
\FOR{$t=T, T-1, ..., 1$}
    \STATE $a\leftarrow \pi(s)$
    \STATE $s', d \leftarrow e.\text{step}(a)$
    \IF{$d$}
        \STATE $s \leftarrow e.\text{reset()}$
    \ENDIF
    \STATE $r \leftarrow \{\Psi^i\}_{i=1}^{N_c}(s)$
    \STATE $\mathcal{D}^{env} \leftarrow \mathcal{D}^{env} \cup \{s, a, r, s', d\}$
    \IF{$t\ \%\ t_{cls\_int} == 0$}
        \FOR{$t'=t_{cls\_steps}, t_{cls\_steps}-1, ..., 1$}
            \STATE $\text{postive samples $\mathcal{D}^{pos}$} \leftarrow \text{sample}(\mathcal{D}^{edit}, \mathcal{B})$
            \STATE $\text{recent negative samples $\mathcal{D}^{neg}$} \leftarrow \text{sample}(\mathcal{D}^{env}_\lambda, \mathcal{B})$
            \STATE $\mathcal{D} \leftarrow \text{mixup}((\mathcal{D}^{pos}, +1),\ (\mathcal{D}^{neg}, -1),\ \alpha)$
            \STATE train classifiers $\{\Psi^i\}_{i=1}^{N_c}$ on $\mathcal{D}$
        \ENDFOR
    \ENDIF
    \STATE one-step training of $\pi, Q_1, Q_2$ follow Yarats et.al.~\cite{yarats2021mastering}
\ENDFOR
\end{algorithmic}
\end{algorithm}

\section{Implementation details}
\label{app: implementation_details}

\subsection{Environment settings}
\label{app: environment_settings}
\paragraph{Wipe environment.} For the wipe simulation environment~(Wipe Sim), we adopt the same setting from the official Robosuite Benchmark~\cite{robosuite2020}. In this task, the agent is expected to clean the markings made up of 100 dark particles on a white table. A dense reward is given based on the distance of the wiper to the stains, the contact between the wiper and table, the particles being wiped, and the completion of the task. A penalty will also be given for collision or joint limit reached, large force, and large acceleration of the wiper. Since this dense reward cannot accurately represent how far along the task is completed, we also record the total number of particles being wiped in an episode and use that value for numerical results in Table~\ref{table:success_rate}.

\paragraph{Push environment.} 
Based on the implementation of the Push task on other environments~\cite{gym-fetch}, we set up a ``Push'' environment in Robosuite. The target of this task is to use the robot arm and end-effector to push a red cube to a specific position. In the visual observations of this environment, the push target is indicated with a green dot, which is at the bottom of the table. The reward of this environment consists of three parts: a dense reward for the reduction of L2 distance between the gripper and the cube, another dense reward for the reduction of L2 distance between the cube and the target, and a one-time reward of 10 points for successfully pushing the cube to the target.

\paragraph{LED environment.} We develop a customized LED environment based on Robosuite, where the agent is expected to turn the color of the light from red to green. A cylinder indicating the LED light is placed at the bottom-right corner of the table, while a button is placed at the bottom-left corner of the table. Once the gripper touches the button or the LED light itself, the color of the LED will change from red to green and the task is considered finished.
The reward of this environment consists of two parts: a dense reward corresponding to the reduction of L2 distance between the gripper and the button, and a completion reward of 10 points for successfully reaching the button or the cylinder. 

\subsection{Reward curve visualization}
For the more challenging real-world scenarios, we visualize the reward function obtained with \ours on several successful trajectories. We first train the reward classifier using 60-80 raw observations as negative samples, and 60-80 edited goal images from these raw observations as positive samples. The training takes 200 epochs, with a batch size of 32 and a learning rate of 1e-4 using Adam as the optimizer. To evaluate these reward classifiers, we collect 5 trajectories for each task and use the trained classifiers to label each frame with a reward score. We also apply normalization on the obtained reward to let each reward curve start from zero at the first frame. We believe this practice enables a fair comparison between different reward functions, as it is the relative trend of the reward curve, rather than the absolute value, that is essential to the downstream RL training. We report the mean and the $95\%$ confidence interval of the reward curve for each task over the collected trajectories.

\subsection{Hyper-parameters for goal image generation}
\label{app: hyper_goal_image}
We provide the hyper-parameters for appearance-based image editing tasks: Wipe~(Sim), Wipe~(Real), LED~(Sim), and LED~(Real) in Table~\ref{table: hyper-appearance-based}, and the hyper-parameters for structure-based image editing tasks: Push~(Sim) and Push~(Real) in Table~\ref{table: hyper-structure-based}. For tasks that require the use of the feature-extracting module in \ours, we provide the additional implementation details of DreamBooth in Table~\ref{table: hyper-dreambooth}.

\begin{table}[t]
  \centering
  \begin{tabularx}{\textwidth}{>{\hsize=.6\hsize}X>{\hsize=.8\hsize}X>{\hsize=.8\hsize}X>{\hsize=.6\hsize}X>{\hsize=.6\hsize}X>{\hsize=.6\hsize}X}
    \toprule              
    Task     & Source Prompt    & Target Prompt &  Cross-Attention Editing Steps & Self-Attention Editing Steps & DreamBooth\\
    \midrule
    Wipe (Sim) & A robot white table with markings on it  & A robot white table with nothing on it & 40 & 0 & Yes \\
    Wipe (Real) & A robot white table with markings on it  & A robot white table with nothing on it & 40 & 0 & Yes  \\
    LED (Sim) & A red cylinder on a white table & A green cylinder on a white table & 40 & 50 & No \\
    LED (Real) & A white table with a yellow duck light &  A white table with a dark duck light & 40 & 0 & No \\
    \bottomrule
  \end{tabularx}
  \caption{\textbf{Hyper-parameters for appearance-based visual goal generation.} The source and target prompt, as well as the editing steps are reported for appearance-based editing tasks. We also report the use of DreamBooth in the feature-extracting module of \ours, please refer to Table~\ref{table: hyper-dreambooth} for details.}
  \label{table: hyper-appearance-based}
\end{table}

\begin{table}[t]
  \centering
  \begin{tabularx}{\textwidth}{>{\hsize=.6\hsize}X>{\hsize=.9\hsize}X>{\hsize=.9\hsize}X>{\hsize=.8\hsize}X>{\hsize=.4\hsize}X>{\hsize=.4\hsize}X}
    \toprule              
    Task     & Prompt    & Bounding-Box &  Tokens & Editing Steps & DreamBooth \\
    \midrule
    Push (Sim) & A photo of a sks cube and a gripper on a white table & [0.5, 0.7, 0.7, 0.9] & $\{5,6,15,...,44\}$ & 10 & Yes \\
    Push (Real) & A photo of a sks cube and a gripper on a white table & [0.1, 0.3, 0.7, 0.9] & $\{5,6,15,...,49\}$ & 10 & Yes \\
    \bottomrule
  \end{tabularx}
  \caption{\textbf{Hyper-parameters for structure-based visual goal generation.} The prompt, bounding box, token set, and the number of editing steps used in the P2P-DD algorithm. We also report the use of DreamBooth in the feature-extracting module of \ours, please refer to Table~\ref{table: hyper-dreambooth} for details.}
  \label{table: hyper-structure-based}
\end{table}

\begin{table}[t]
  \centering
  \begin{tabularx}{\textwidth}{>{\hsize=.6\hsize}X>{\hsize=.6\hsize}X>{\hsize=1.6\hsize}X>{\hsize=.4\hsize}X>{\hsize=.4\hsize}X>{\hsize=.4\hsize}X}
    \toprule              
    Task     & Instance Number    & Class Prompt &  Class Images & Training Steps & Learning Rate \\
    \midrule
    Wipe (Sim) & 20  & a photo of a Franka robot gripper & 200 & 800 &  5e-6 \\
    Wipe (Real) & 12  & a photo of a Franka robot gripper & 200 & 800 &  5e-6  \\
    Push (Sim) & 11 & a photo of a red cube & 200 & 800 & 5e-6 \\
    Push (Real) & 8 & a photo of a red cube & 200 & 800  & 5e-6 \\
    \bottomrule
  \end{tabularx}
  \caption{\textbf{Hyper-parameters of the DreamBooth technique used in the feature-extracting module of \ours.} The training details of DreamBooth for each task. DreamBooth extracts a special token that preserves the details of an object in the scene.
  }
  \label{table: hyper-dreambooth}
\end{table}

\subsection{Hyper-parameters for example-based RL}
\label{app: hyper_rl}

We provide the training details of the example-based RL algorithm in \ours in Table~\ref{table: hyper-rl}.
Unless otherwise specified, we use the same hyperparameters as those in DrQv2. For each curve in the paper, we run 3 seeds and report the mean and 95\% confidence interval of the results.
\begin{table}[!h]
    \centering
    \begin{tabular}{lc}
        \toprule
        Hyperparameter & Value \\
        \midrule
        Batch Size & 128 \\
        Mixup alpha & 1 \\
        Classifier Layer Sizes & [1024, 2] \\
        Classifier Ensemble & 10 \\
        Negative Sample Ratio $\alpha$ & 0.05\\
        Classifier Update Interval & 1000 \\
        Classifier Update Steps & 10 \\
        Edited Dataset Size & 1024 \\
        Total Train Steps & 10e6 \\
        \bottomrule
    \end{tabular}
    \caption{Example Based RL Hyper-Parameters}
  \label{table: hyper-rl}
\end{table}

\subsection{User study}
\label{app: user study}

\begin{figure}[t]
    \centering
    \includegraphics[width=\linewidth]{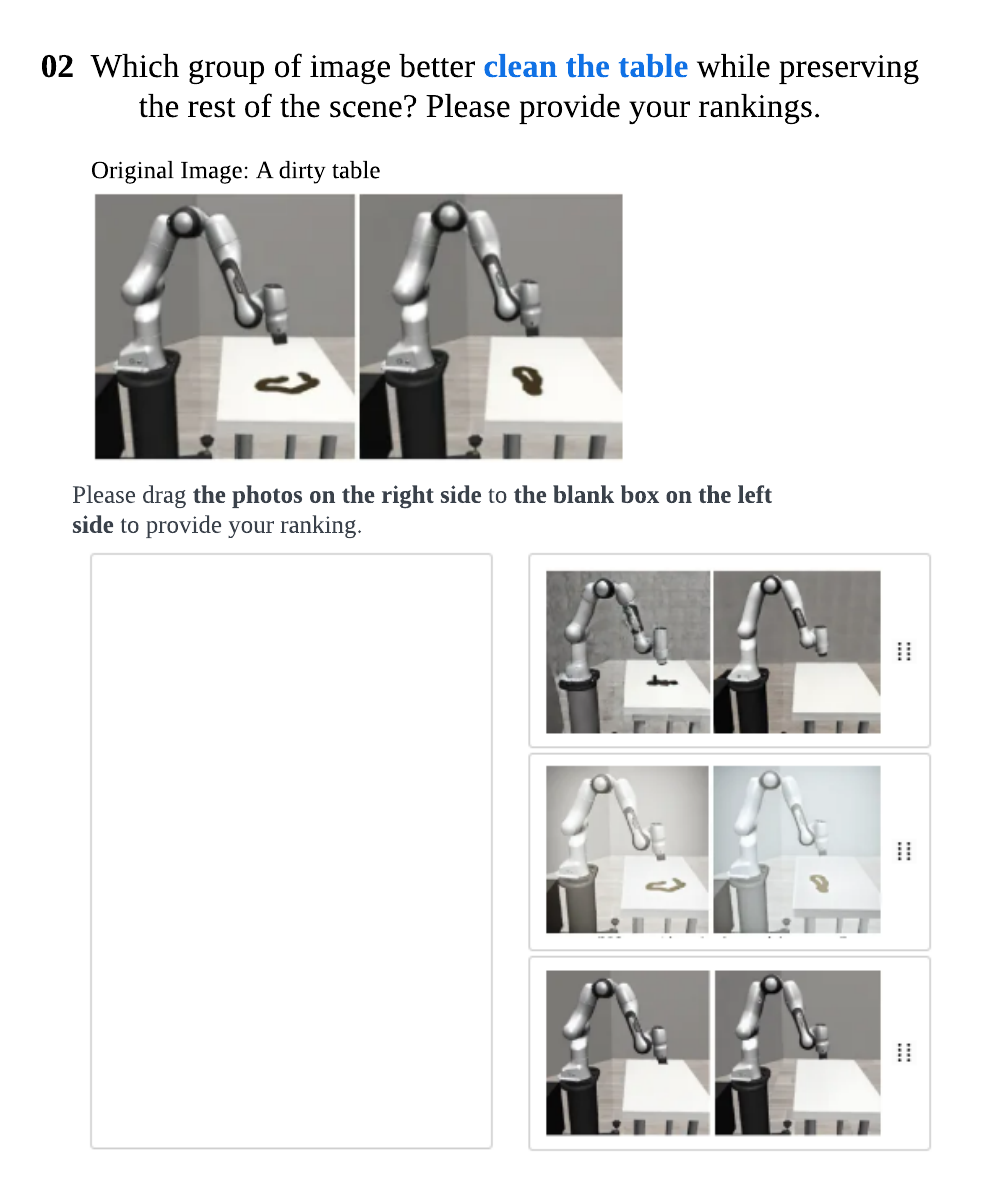}
    \caption{\textbf{A print screen for the questionnaire used in the user study.}}
    \label{fig:user-study}
\end{figure}

For the human user study on visual goal generation,
We include 24 random subjects who do not have specialized knowledge in artificial intelligence. The subjects are shown images in Figure~\ref{fig:goal_generation_1} and Figure~\ref{fig:goal_generation_2}, together with the editing instructions, and are then asked to provide rankings of each editing method based on how well the editing instruction is followed in each result. 
We also provide a snapshot of the questionnaire sent to the subjects in Figure~\ref{fig:user-study}.

We utilize the Glicko-2 rating system to analyze the ranking results. The Glicko-2 rating is an estimation of a candidate's strength based on its performance relative to others. To utilize this system, we first transform the rankings obtained in the questionnaire into comparisons: if method $m_1$ is ranked higher than method $m_2$, then $m_1$ is marked as a victory in the comparison between $m_1$ and $m_2$. Then we determine the rating of each method based on the initial score and the win-loss information contained in these comparisons, as suggested in the Glicko-2 system~\cite{glickman2013glicko}. Finally, to standardize the scores within a scale from 0 to 100, we use the following formula to perform normalization:
\[R_{normal}=\frac{R-800}{2000-800}*100\]
 
\section{Additional results}
\label{app: additional_results}

\begin{figure}[h!]
    \centering
    \includegraphics[width=\linewidth]{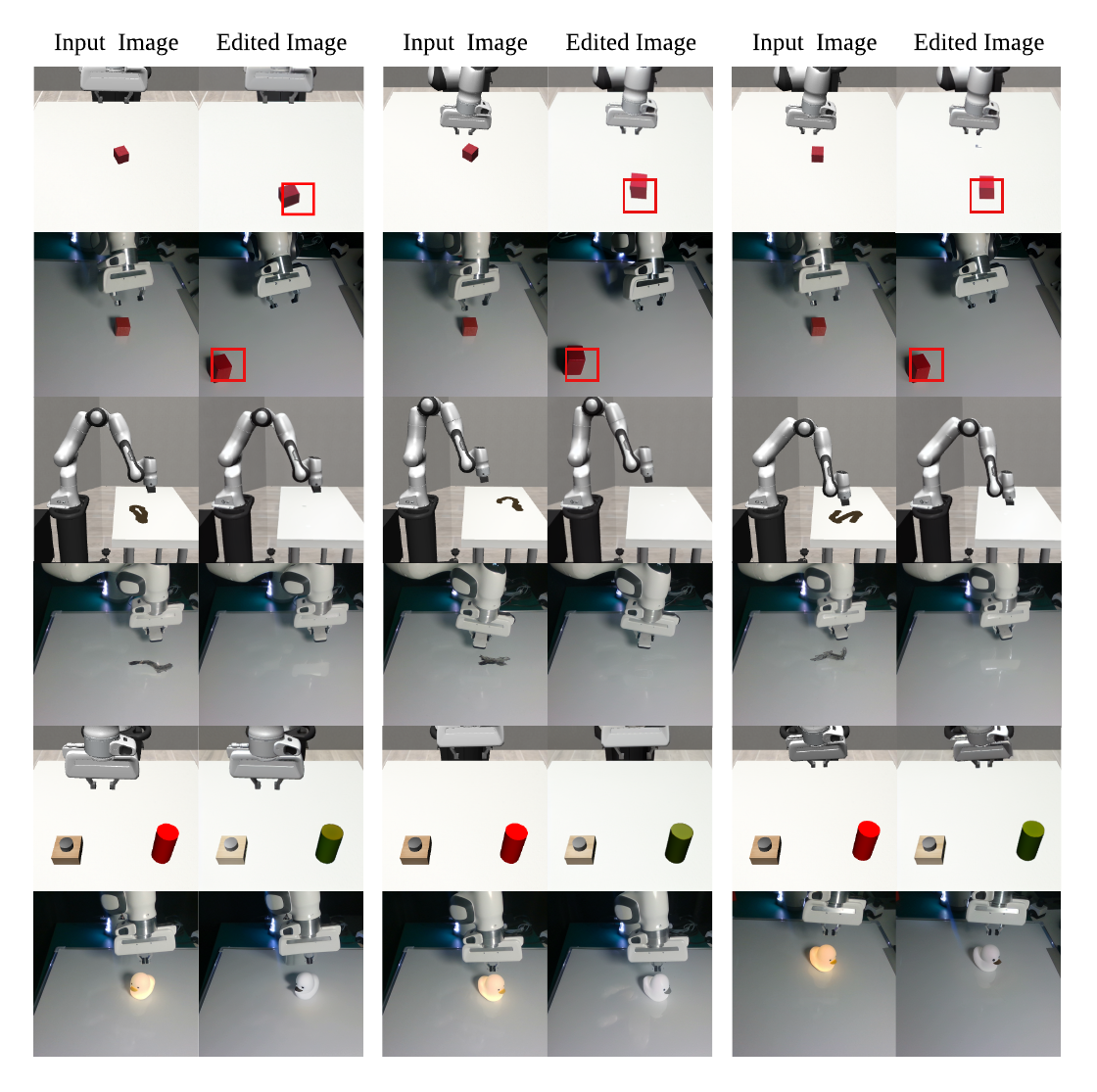}
    \caption{\textbf{Additional goal generation results.} We demonstrate several additional visual goal generation examples of \ours. From the top row to the bottom are the results of: Push~(Sim), Push~(Real), Wipe~(Sim), Wipe~(Real), LED~(Sim), and LED~(Real).} 
    \label{fig:additional-goal-generation}
\end{figure}

\paragraph{Visual goal generation.} We provide additional visual goal-generation results for each task in Figure~\ref{fig:additional-goal-generation}. The examples are selected from the datasets generated by \ours that are used to train example-based RL algorithms or reward functions for real-world evaluation.

\paragraph{Visualization of reward functions.} We also provide visualization videos of the reward functions obtained through \ours evaluated on both successful and failure trajectories. Please visit: \href{https://lfvoid-rl.github.io/}{\texttt{LfVoid.github.io}}.

\section{Limitations}
\label{app: limitation}

Despite \ours's successful performance in editing images, it still has several limitations. We summarize the common failure cases into the following four categories: object deformations, background alterations, size changes, and incorrect associations. We will further discuss the causes of each failure case and suggestions to avoid these failures in the following paragraphs.

\paragraph{Object deformations.} Objects might undergo deformations, including shape changes and color loss, between the input and edited images, as shown in Figure~\ref{fig:object-deformation}. Hyper-parameter tuning can alleviate this problem by finding the best editing step for each input image, and the optimal value usually is the same for different input images of the same scene. But the optimal parameters for various scenes are usually different, thus requiring extra parameter tuning.

\paragraph{Background alterations.} In some cases, details of the background scene may undergo unintended changes during the editing process, as shown in Figure~\ref{fig: background-changes}. The DreamBooth technique can mitigate this problem by fine-tuning the diffusion model on several images of the background to learn its details and structures. Performing grid-search on the hyper-parameter used in the editing techniques can also alleviate background alterations. In some cases, prompt engineering can also be a solution to this problem by adding additional tokens that describe the background to better preserve its details.

\paragraph{Size changes.} The use of Directed Diffusion may alter the size of the object, typically making it larger than its original size, as illustrated in Figure~\ref{fig: size-changes}. This phenomenon is the result of the imprecise control used in Directed Diffusion, which applies attention strengthening on the whole bounding-box area, rather than an exact subset of the area where we want the object to be placed. Moreover, the large-scale pre-trained diffusion models do not seem to possess the ability to change the shape of objects according to the viewpoint. Training diffusion models with knowledge of the physical laws may be needed to solve this problem.

\paragraph{Incorrect associations.} There have been instances where the model fails to associate objects with the corresponding prompt word, as shown in the examples in Figure~\ref{fig: incorrect-association}. Pre-trained diffusion models sometimes fail to correctly associate the input image with the given prompts, thus failing to follow the editing instruction. One solution to mitigate this problem is to use prompt engineering to select the best prompt that the model can "understand": the attention maps can precisely highlight the part of the image corresponding to each token. Training diffusion models with a better understanding of natural language and images may be the ultimate solution to this problem.

\begin{figure}[t]
\begin{subfigure}{\linewidth}
  \centering
  \includegraphics[width=\linewidth]{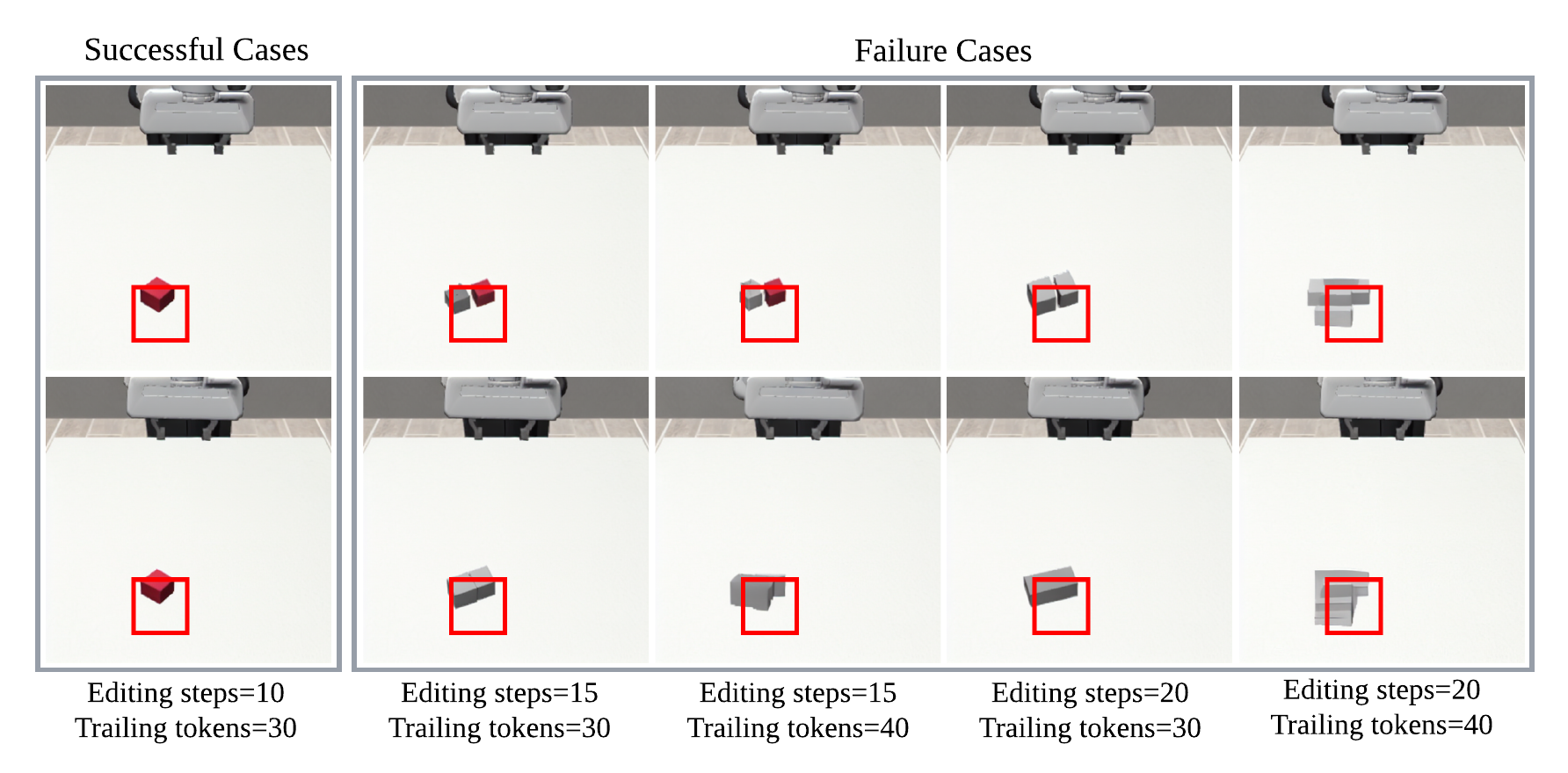}
  \caption{\textbf{Failure cases: object deformations.} Different hyper-parameters may result in failure editing results that change the object's shape and color.}\label{fig:object-deformation}
\end{subfigure} 

\begin{subfigure}{\linewidth}
\includegraphics[width=\linewidth]{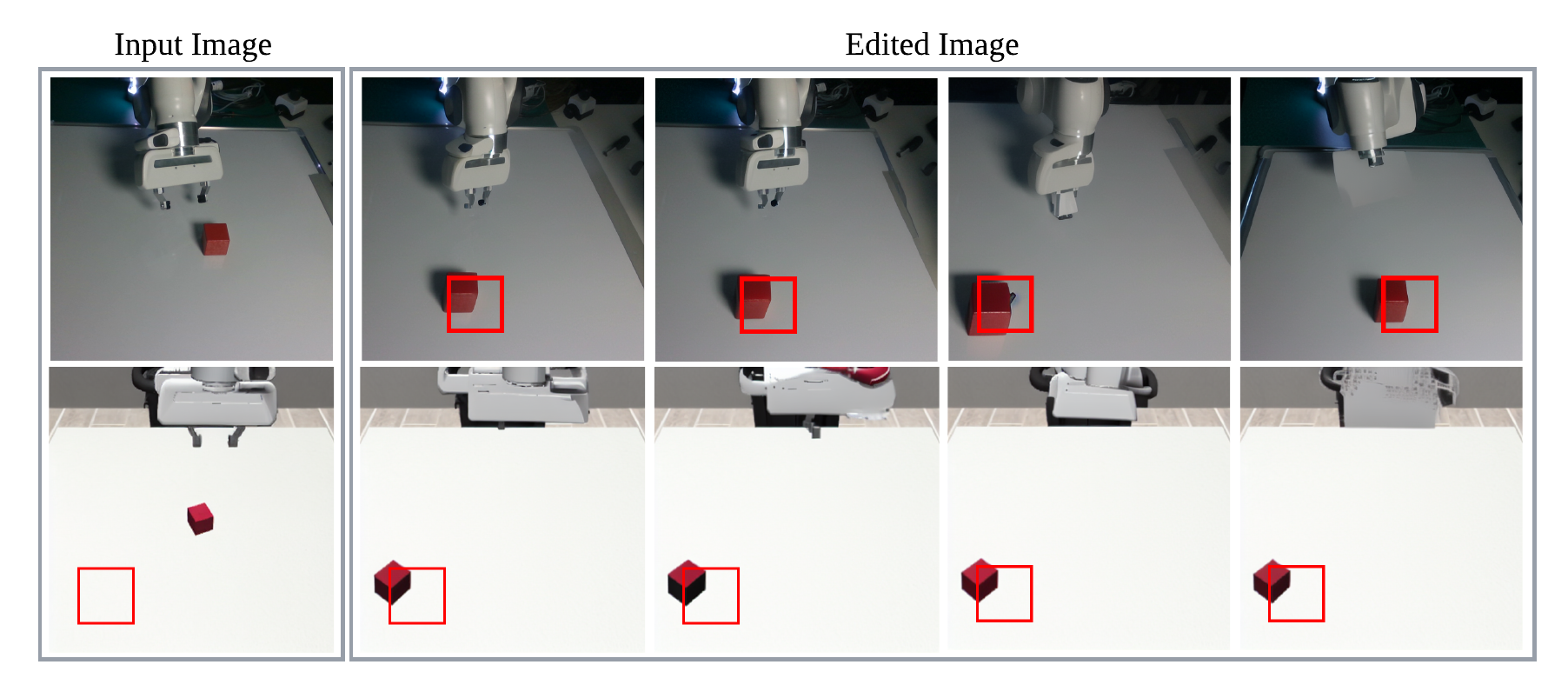}
\caption{\textbf{Failure cases: background alterations.} Details of the background may change during the editing process. The DreamBooth technique, performing grid-search on hyper-parameters, as well as prompt engineering can mitigate this problem.
}
\label{fig: background-changes}
\end{subfigure}
\hfill

\begin{subfigure}{\linewidth}
\includegraphics[width=\linewidth]{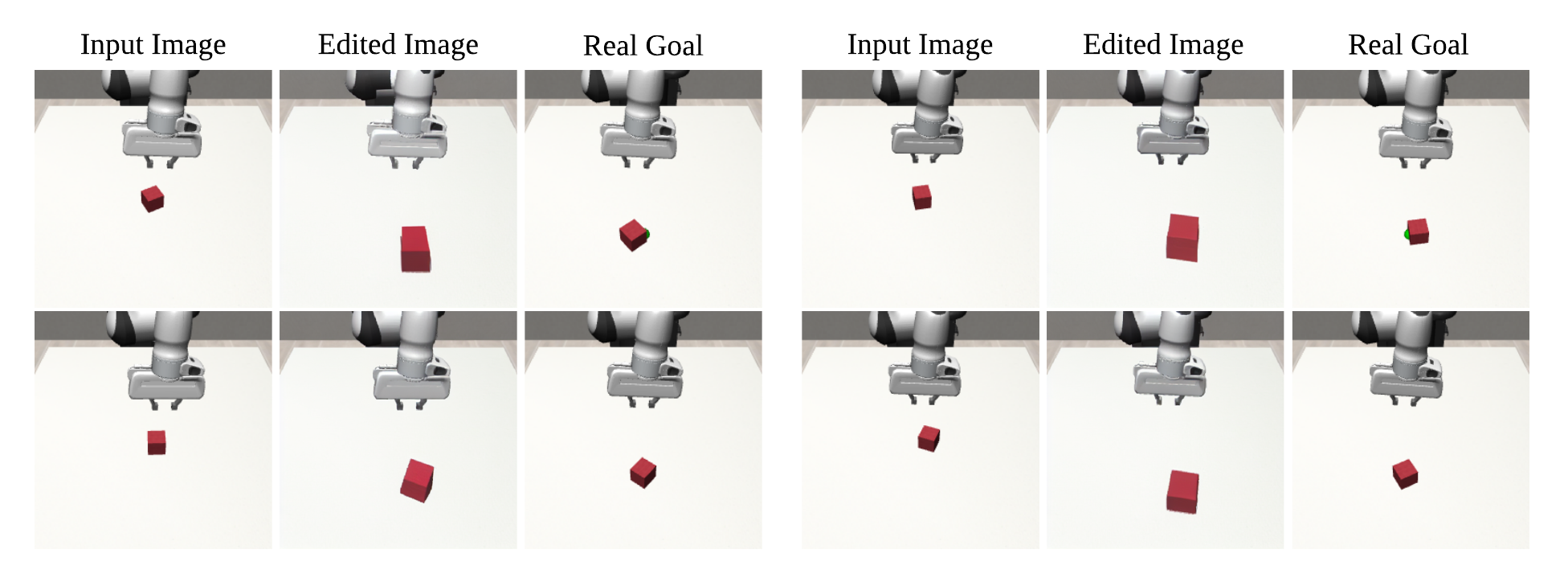}
\caption{\textbf{Failure cases: size changes.} The Directed Diffusion technique may cause the diffusion model to generate images with objects larger than the original ones.}
\label{fig: size-changes}
\end{subfigure}
\hfill

\end{figure}

\begin{figure}[t]\ContinuedFloat
\begin{subfigure}{\linewidth}
\includegraphics[width=\linewidth]{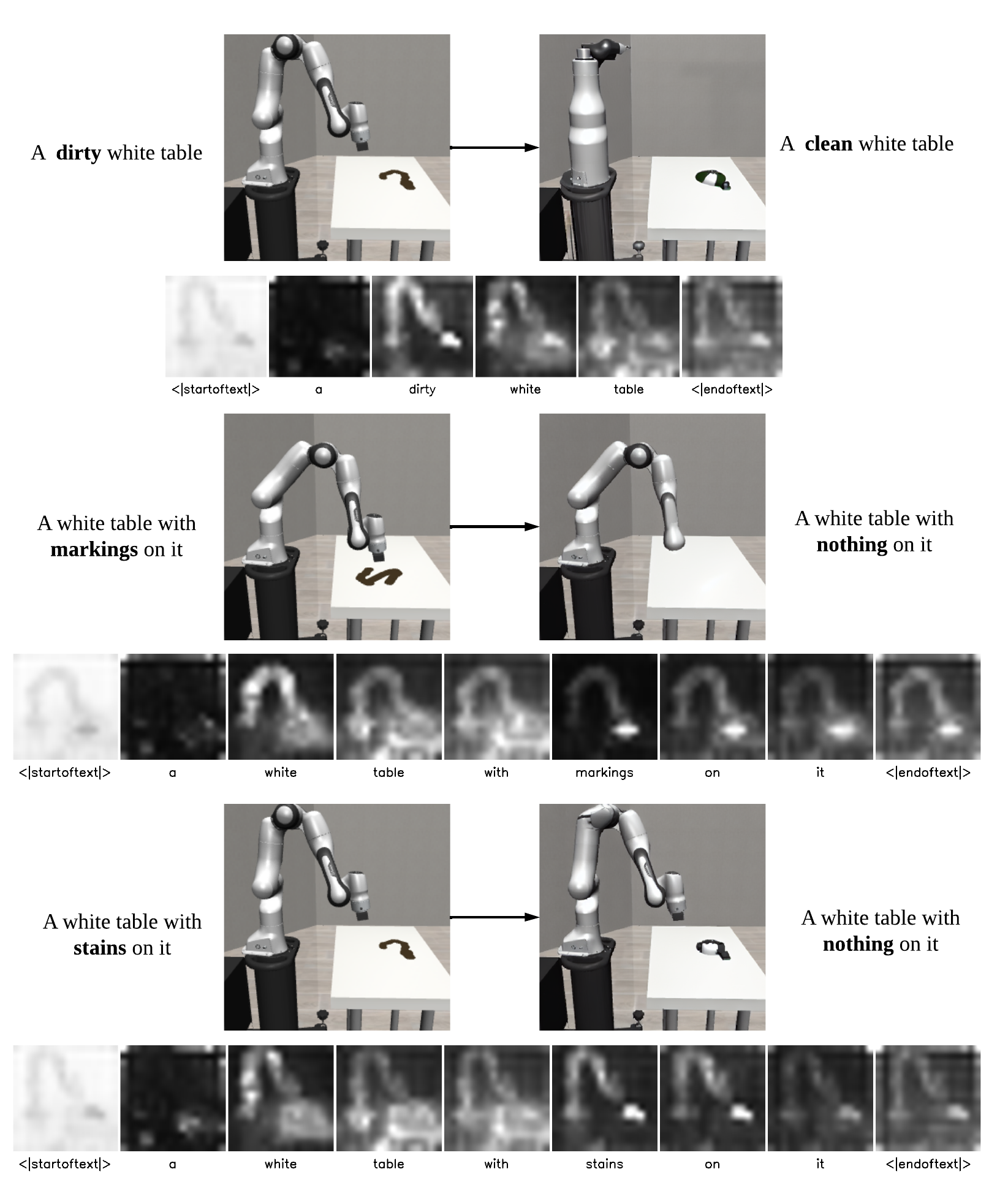}
\caption{\textbf{Failure cases: incorrect associations.} Different prompts may result in different editing qualities depending on how well the model can associate each token with its corresponding area in the input image.}
\label{fig: incorrect-association}
\end{subfigure}
\hfill
\caption{\textbf{Failure cases of visual goal generation.} We demonstrate four common failure cases when using \ours to generate visual goals in various scenes: object deformations, background alterations, size changes, and incorrect associations.}
\end{figure}